\documentclass[journal]{IEEEtran}
\usepackage{amsmath,amsfonts}
\usepackage{algorithmic}
\usepackage{algorithm}
\usepackage{array}
\usepackage[caption=false,font=normalsize,labelfont=sf,textfont=sf]{subfig}
\usepackage{textcomp}
\usepackage{stfloats}
\usepackage{circledsteps}
\usepackage{url}
\usepackage{verbatim}
\usepackage{graphicx}
\usepackage{tabularx}
\usepackage[table]{xcolor}
\usepackage{cite}
\usepackage{mathtools}
\usepackage{booktabs} 
\usepackage{ragged2e}
\usepackage{pifont}
\newcolumntype{L}[1]{>{\RaggedRight\arraybackslash}m{#1}}
\newcolumntype{Y}{>{\raggedright\arraybackslash}X}
\usepackage{pifont}
\usepackage[pagebackref=false,colorlinks=true,linkcolor=blue,citecolor=blue,urlcolor=blue]{hyperref}
\usepackage[capitalize]{cleveref}

\usepackage{etoolbox}
\makeatletter
\pretocmd{\@maketitle}{%
  \noindent\begin{minipage}{\textwidth}\normalsize\rmfamily\raggedright
  This paper is accepted and will appear in the IEEE Transactions on Robotics.
  \copyright 2026 IEEE. Personal use of this material is permitted. Permission from IEEE must be
  obtained for all other uses, in any current or future media, including reprinting/republishing
  this material for advertising or promotional purposes, creating new collective works, for resale
  or redistribution to servers or lists, or reuse of any copyrighted component of this work in
  other works.
  \end{minipage}\par\vspace{1.5\baselineskip}%
}{}{\PackageWarning{DexRemo}{Could not place the IEEE notice above the title}}
\makeatother

\begin{document}

\title{DexRemo: A Force-Sensing Hand Platform and Task Suite for Contact-Rich Removal Manipulation}

\author{Liman Wang$^{1,\dagger}$, Hanyang Zhong$^{1,\dagger}$, Sun Zhaole$^{2}$, Wenting Cao$^{3}$,\\
Babar Jamil$^{1}$, Andy Tyrrell$^{1}$, Shan Luo$^{4}$, and Jihong Zhu$^{1,*}$%
\thanks{$^{\dagger}$Equal contribution. $^{*}$Corresponding author.}%
\thanks{$^{1}$Liman Wang, Hanyang Zhong, Babar Jamil, Andy Tyrrell, and Jihong Zhu are with the
School of Physics, Engineering and Technology, University of York, York, U.K.
(e-mail: liman.wang@york.ac.uk; hanyang.zhong@york.ac.uk; jihong.zhu@york.ac.uk).}%
\thanks{$^{2}$Sun Zhaole is with the University of Edinburgh, Edinburgh, U.K.}%
\thanks{$^{3}$Wenting Cao is with Friedrich Schiller University Jena, Jena, Germany.}%
\thanks{$^{4}$Shan Luo is with the Department of Engineering, King's College London, London, U.K.}}

\markboth{}{}

\maketitle

\begin{abstract}
Many manipulation tasks require separating adhered, embedded, contaminated, or mechanically constrained targets. We study this setting as \emph{removal manipulation}, where progress depends on evolving contact states that are only partially observable from vision and kinematics. We introduce \emph{DexRemo}, a three-finger robotic hand with high-force actuation, modular fingertips, and three-axis fingertip force sensing, together with a nine-scenario real-robot task suite spanning adhesive detachment, surface cleaning and self-cleaning, and occluded retrieval. To make multi-finger force observations easier to fuse, we express fingertip forces in a shared wrist-centered frame, giving force components consistent geometric meaning across fingers and time. Using an Action Chunking Transformer backbone, policies fuse multi-view vision, proprioception, fingertip pose, and force observations to predict closed-loop action chunks from demonstrations. Evaluation uses six core benchmark tasks for controlled fingertip-input and force-frame ablations, plus three extension scenarios for additional held-out coverage. Across these evaluations, the learned policies achieve consistent removal performance, and ablations show that wrist-frame forces, especially with fingertip pose, improve stability and generalization over local-frame force inputs and no-force baselines. Project website:
\url{https://dexremo.github.io}.
\end{abstract}

\section{Introduction}

\begin{figure}[!t]
\centering
\includegraphics[width=1\linewidth]{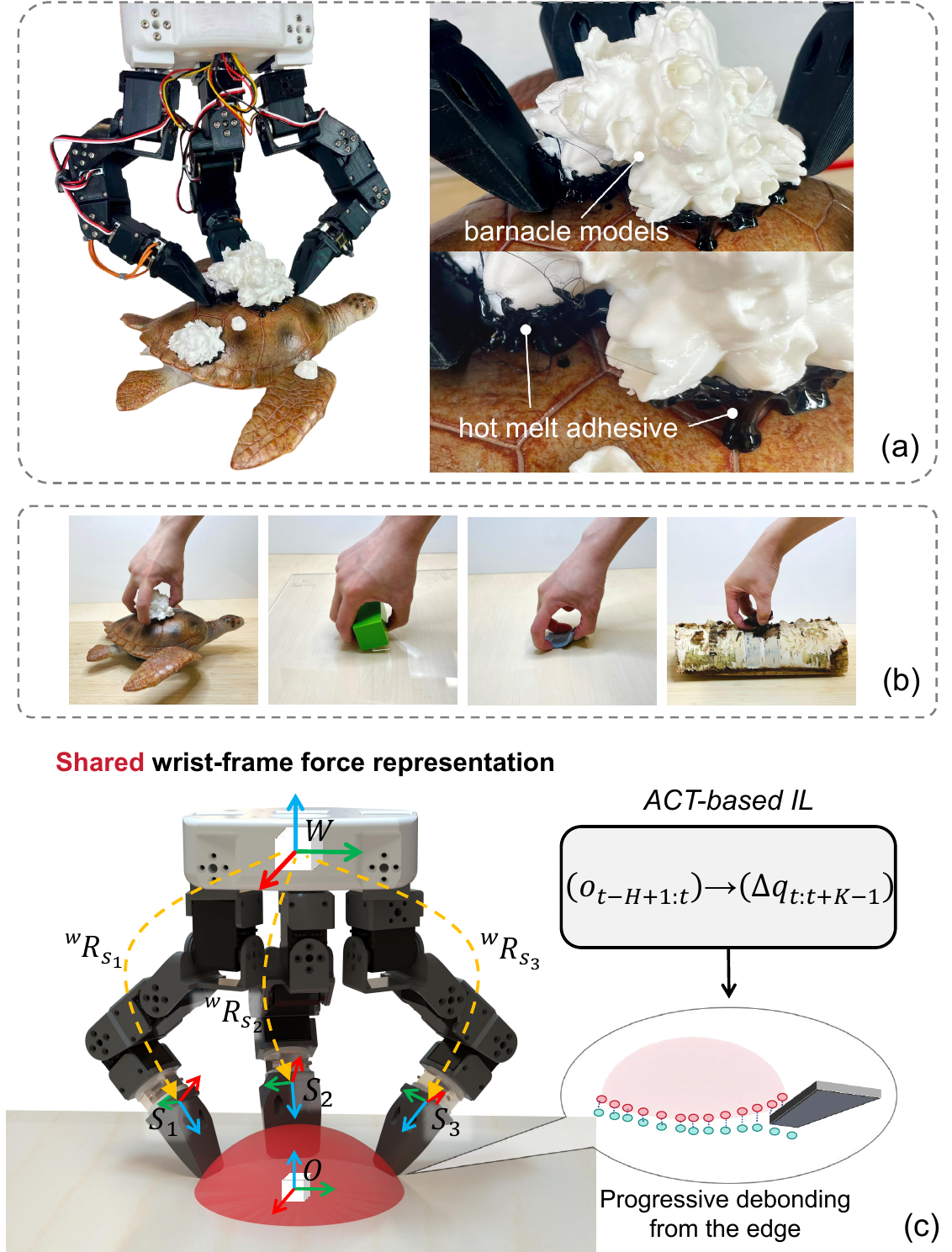}
\caption{\textbf{Overview of removal manipulation and the proposed framework.}
(a) DexRemo Hand removes strongly adhered 3D-printed barnacles from a curved turtle mockup, demonstrating variable and partially observable attachments.
(b) Representative human removal behaviors reveal task-intrinsic interaction principles.
(c) Fingertip forces are transformed from local sensor frames ${\mathbf{R}_{S_i}}$ to a shared wrist frame, then fused with vision, fingertip poses, and proprioception. An ACT-based policy generates observation-conditioned action chunks for progressive, edge-initiated debonding.}
\label{fig_1}
\end{figure}

\IEEEPARstart{R}{obotic} manipulation in unstructured environments often involves targets that cannot be directly manipulated because they are adhered, embedded, contaminated, or constrained by surrounding materials. Such tasks require prior separation before conventional manipulation can proceed. We term this class of problems \emph{removal manipulation}: tasks in which a robot progressively separates, exposes, or cleans a target through contact-rich behaviors such as detachment, peeling, scraping, prying, wiping, or uncovering, as illustrated by representative human examples in Fig.~\ref{fig_1}(b). Success in these tasks depends not only on object geometry, but also on evolving physical constraints such as adhesion, frictional wedging, deformation, occlusion, and surface contamination.
Removal manipulation is challenging because progress depends on the evolving contact state between the target and its surroundings. This state is only partially observable from vision and kinematics: visually similar scenes may correspond to different adhesion, friction, deformation, or occlusion conditions, while small contact changes can determine whether the robot initiates separation, maintains contact, or damages nearby material. Successful behavior therefore requires force-capable interaction, geometrically consistent multi-finger contact feedback, and temporally adaptive policies that revise actions as contacts form, slide, stick, and break.

Prior work on contact-rich separation has mainly addressed specific tasks, including robotic peeling~\cite{MORPHeus}, compliant surface cleaning~\cite{Scraping}, and tactile-driven excavation or retrieval in deformable or particulate media~\cite{Tactile_Excavation,Object_Retrieval}. Tactile sensing has also supported slip detection, contact geometry inference, and compliant surface following~\cite{yamaguchi2019leveraging,wang2021gelsightwedge}. Although these studies demonstrate the value of contact feedback, removal behaviors have not been systematically evaluated in a common real-robot setting that combines force-capable multi-finger hardware, task-level physical success criteria, and tasks involving adhesion, contamination, burial, and occlusion.

To address this gap, we introduce \emph{DexRemo}, a three-finger robotic hand with three-axis fingertip force sensing, together with a real-robot task suite spanning adhesive detachment, surface cleaning and self-cleaning, and occluded retrieval. The suite captures recurring contact-rich separation challenges under different physical constraints, with explicit preparation procedures, physical success criteria, and held-out variations in object geometry, pose, color, and attachment conditions. Table~\ref{tab:pullout} reports pullout strength and power density as task-relevant capability indicators. Across tasks, peak fingertip loads frequently reach tens of newtons (Sec.~\ref{sec:platform_validation}), reflecting the sustained contact required by the evaluation criteria.

A central design issue is how to represent multi-finger force measurements for learning. Each fingertip sensor measures forces in its local frame, which rotates with finger articulation. As a result, the meaning of each local force component varies with fingertip pose and across fingers, complicating cross-finger fusion. We therefore transform all fingertip forces, using forward kinematics, into a shared wrist-centered frame with consistent geometric meaning across fingers and time (Fig.~\ref{fig_1}). This deterministic transformation follows classical task-frame representations; our goal is to evaluate its practical value for multimodal imitation learning with 20 demonstrations per task and articulated fingertip force sensors.

For policy learning, we use an Action Chunking Transformer (ACT)~\cite{Zhao2023ACT} as the multimodal sequence-prediction backbone. The policy fuses multi-view vision, proprioception, fingertip pose, and fingertip force observations, and predicts short-horizon joint action chunks in a receding-horizon loop. ACT provides a consistent temporal imitation-learning backbone for studying how force sensing and force representation affect contact-rich removal performance.

Under the two-tier evaluation protocol, task-specific policies achieve consistent held-out performance across the suite. Controlled ablations on the six core benchmark tasks show that combining fingertip pose with wrist-frame force sensing improves stability and generalization over local-frame force inputs, no-force baselines, and other evaluated fingertip-input configurations. These results identify shared-frame force alignment as a practical representation choice for multimodal learning in the evaluated setting of 20 demonstrations per task.

The contributions of this work are:
\begin{itemize}
\item \textbf{A force-sensing hand platform and real-robot removal task suite.} We introduce \emph{DexRemo}, a reproducible three-finger robotic hand with high-force actuation, modular fingertips, and integrated three-axis fingertip force sensing, together with a nine-scenario suite covering adhesive detachment, surface cleaning and self-cleaning, and occluded retrieval. The hand reaches a measured pullout strength of $62.5$~N and a power density of $3.064$, the highest among the hands compared in Table~\ref{tab:pullout}, at a hand cost of about \$200 plus \$170 per optional fingertip sensing unit (Table~\ref{tab:multi_fingered_hands}). The suite defines preparation procedures, physical success criteria, held-out tests, and a two-tier evaluation protocol.
\item \textbf{A controlled empirical study of shared-frame force alignment for multi-finger imitation learning.} We represent fingertip forces in a wrist-centered frame grounded in classical task-frame formulations, and compare wrist-frame, local-frame, and no-force inputs while holding the hand, demonstrations, policy backbone, training schedule, and rollout protocol fixed. Real-robot ablations show that explicit force alignment, especially when combined with fingertip pose, yields the strongest and most consistent performance among the evaluated configurations.
\end{itemize}

\begin{table}[h]
\centering
\small
\caption{\textbf{Pullout strength and power density comparison.} Peak pullout force and corresponding power density for representative hands. Metrics are defined in Appendix~\ref{app:pullout_protocol}.}
\resizebox{\columnwidth}{!}{  
\begin{tabular}{|l|c|c|}
\hline
\textbf{Hand} & \textbf{Strength (N)} & \textbf{Power Density $N \times DOF/(\text{cm}^2)$} \\
\hline
Bauer et al. \cite{bauer}      & 37.4  & 0.677 \\
Allegro Hand \cite{allegro}    & 8.5   & 0.350 \\
D'Manus \cite{dmanus}          & 27.8  & 0.313 \\
Inmoov Hand \cite{inmoov}      & 5.8   & 0.116 \\
LEAP Hand \cite{shaw2023leap}          & 19.5  & {1.045} \\
LEAP-C Hand \cite{shaw2023leap}        & 21.5  & {1.15} \\
Adult Human Hand \cite{shaw2023leap}             & 26.5  & 2.199 \\
\textbf{DexRemo Hand (Ours)}      & \textbf{62.5} & \textbf{3.064} \\
\hline
\end{tabular}
}
\label{tab:pullout}
\end{table}

\begin{table}[t]
\centering
\caption{Representative multi-finger hands/platforms: finger count, transmission, approximate
cost (USD), and actuated DoFs. Costs may vary by configuration. DL: direct linkage; T: tendon. Note: Costs are reported for the hands/platforms only and exclude any tactile/force sensing modules.}
\setlength{\tabcolsep}{1mm}{
\begin{tabular}{|lcccc|}
\hline
\textbf{Hands / Platforms} & \textbf{Finger Num.} & \textbf{Transmission} & \textbf{Cost (USD)} & \textbf{DoFs} \\
\hline
Allegro Hand-4F~\cite{allegro}     & 4 & DL & 16,000 & 16 \\
LEAP Hand~\cite{shaw2023leap}              & 4 & DL & 2,000  & 16 \\
Delta Hand~\cite{delta}            & 4 & DL & 1,000  & 12 \\
Shadow Hand~\cite{shadow}          & 5 & T  & 300,000 & 20 \\
Inspire Hand~\cite{inspire}        & 5 & T  & 5,000   & 6  \\
Ability Hand~\cite{AbilityHand}    & 5 & T  & 20,000  & 6  \\
Tesollo Hand-5F~\cite{tesllo}      & 5 & DL & 16,900  & 20 \\
\hline
Barrett Hand~\cite{barret}         & 3 & T  & 50,000  & 4  \\
EyeSight Hand~\cite{eyesight}      & 3 & DL & 2,500   & 7  \\
Allegro Hand-3F~\cite{allegro}     & 3 & DL & N/A       & 9  \\
Tesollo Hand-3F~\cite{tesllo}      & 3 & DL & 9,500   & 20 \\
D'Claw~\cite{ahn2019robel} & 3 & DL  & 3,500 & 9  \\
DexRemo Hand (Ours)             & 3 & DL & 200 & 12 \\
\hline
\end{tabular}
}
\label{tab:multi_fingered_hands}
\end{table}

The remainder of this paper is organized as follows. Section~\ref{section2} reviews related work. Section~\ref{section3} defines the local removal setting and presents the DexRemo platform, sensing system, and wrist-frame force representation. Section~\ref{section5} presents the multimodal imitation-learning framework. Section~\ref{section6} describes the task suite, experimental protocol, and evaluation results. Section~\ref{section8} discusses limitations, future work, and conclusions.

The hardware design files, simulation models, software, and supplementary videos are being released in stages on the project website.

\section{Background and Related Work}
\label{section2}

We review contact-rich removal manipulation, force-aware and spatially grounded tactile learning, evaluation suites, and dexterous hand platforms, motivating DexRemo as a real-robot system for studying forceful removal with consistent multi-finger force representation and controlled held-out evaluation.

\subsection{Contact-Rich Separation and Structured Evaluation}

Most manipulation pipelines assume objects are directly accessible and separable, focusing on grasping and object-centric control~\cite{bohg2014data,mahler2017dex,fang2020graspnet}. Removal manipulation instead requires progressive separation under adhesion, contamination, embedding, or mechanical constraints, where progress depends on contact feedback and force regulation. Prior work has studied specific separation behaviors, including contact-geometry sensing~\cite{wang2021gelsightwedge}, robot-assisted peeling~\cite{MORPHeus}, force-regulated cleaning~\cite{Scraping}, and tactile-driven excavation or retrieval~\cite{Tactile_Excavation,Object_Retrieval}. These works validate the importance of interaction feedback, but are typically scoped to a single task family or specialized geometry. We complement them with a multi-task real-robot setting spanning adhesive detachment, surface cleaning, self-cleaning, and occluded retrieval under a shared protocol.

Task sets enable systematic comparison by defining tasks, observations, actions, success criteria, and test conditions. Simulation platforms such as ManiSkill3 scale contact-rich evaluation under physics-based contacts~\cite{tao2024maniskill3}. Tactile-focused evaluations such as ManiSkill-ViTac provide unified protocols for vision--tactile manipulation~\cite{li2024maniskillvitac2025}, and VTDexManip introduces a visual--tactile dataset and benchmark for dexterous manipulation~\cite{liu2025vtdexmanip}. These suites are valuable for visual--tactile learning, but primarily target general dexterous skills, simulation-based evaluation, or tactile-vision policy learning. Our suite is complementary: it focuses on real-robot progressive separation under adhesion, contamination, and occlusion, with physical success criteria and held-out variations designed to stress contact-state uncertainty and forceful interaction. Because real-robot evaluation is sample-limited, we also report statistical analyses motivated by recent work on rigorous small-sample policy comparison~\cite{snyder2025policycomparison}.

\subsection{Force Representations and Multimodal Policy Learning}

Contact-rich manipulation has motivated policies that combine force or tactile feedback with vision and proprioception. ForceMimic collects and imitates force--motion demonstrations~\cite{liu2025forcemimic}, ForceVLA fuses real-time six-axis force/torque feedback with visual--language features during action decoding~\cite{yu2025forcevla}, and 3D-ViTac combines visual and tactile observations for fine-grained dexterous manipulation~\cite{huang2024vitac}. These policies also differ in their prediction targets and control interfaces: some predict force or torque as an auxiliary supervised output while retaining positional actions~\cite{zhang2025elucidating}, whereas direct force-control methods provide force or wrench references tracked by a lower-level force or impedance controller~\cite{bruyninckx1996taskframe,salisbury1982articulated}. DexRemo instead uses measured fingertip forces as observations and predicts joint-position displacement chunks, without an auxiliary force-prediction loss or explicit force/wrench commands. This fixed positional-action pipeline allows us to isolate the effect of fingertip-force input representation.

Representing interaction forces in task-relevant frames is a classical principle. Task-frame control expresses force actions in frames aligned with task constraints~\cite{bruyninckx1996taskframe}, and articulated-hand force control builds on the grasp matrix and structured hand--contact models~\cite{salisbury1982articulated}. SaTA~\cite{huang2026sata} computes each vision-based tactile sensor's 6D pose through forward kinematics, Fourier encodes the pose, and uses FiLM to condition tactile-image features. DexRemo instead applies a deterministic coordinate rotation to calibrated three-axis fingertip force vectors and compares local-frame, wrist-frame, and no-force inputs under the same ACT-based pipeline. Thus, SaTA studies a learned spatially conditioned representation for tactile images, whereas our study isolates force-vector coordinate representation in real-robot removal manipulation.

Transformer-based imitation learning supports temporally coherent behavior through action-chunk prediction and observation-conditioned replanning~\cite{Zhao2023ACT}. Contact-aware ACT variants adopt different sensing and control designs: TACT~\cite{murooka2025tact} encodes distributed tactile measurements as a tactile token for humanoid whole-body manipulation; FTACT~\cite{watanabe2025ftact} augments ACT with force/torque, visual, and joint observations; and Bi-ACT~\cite{buamanee2024biact} uses position, velocity, and force information for bilateral position--force control. DexRemo retains a joint-position action interface and represents calibrated fingertip forces and poses with separate MLP tokens. We keep the ACT backbone fixed to isolate force availability and coordinate representation, using action chunks for coordinated multi-finger motion and periodic replanning to condition subsequent actions on updated visual, proprioceptive, fingertip-pose, and force observations.

\subsection{Dexterous Hands and Force Sensing for Contact-Rich Interaction}

Multi-finger hands support dexterous and contact-rich behaviors, including widely used platforms such as Allegro and Shadow~\cite{allegro,shadow}, and open designs such as LEAP Hand~\cite{shaw2023leap} and ORCA~\cite{christoph2025orca}. These platforms have advanced reproducible dexterous manipulation, especially for grasping, in-hand manipulation, teleoperation, and robot learning. Removal manipulation imposes additional requirements: fingertip accessibility, multi-point controllability, sustained normal and tangential load transmission, and force sensing during contact-rich interaction. DexRemo addresses these needs with interchangeable rigid and compliant fingertips, integrated three-axis fingertip force sensing, and a wrist-frame multi-finger force representation. 
We report pullout strength and power density as task-relevant capability indicators in Table~\ref{tab:pullout}, with metric definitions provided in Appendix~\ref{app:pullout_protocol}, and evaluate sensing inputs and force-coordinate representations while holding the hand, demonstrations, policy architecture, training schedule, and rollout protocol fixed. For broader platform context, Table~\ref{tab:multi_fingered_hands} summarizes representative platforms.

\begin{figure*}[!t]
\centering
\includegraphics[width=1\linewidth]{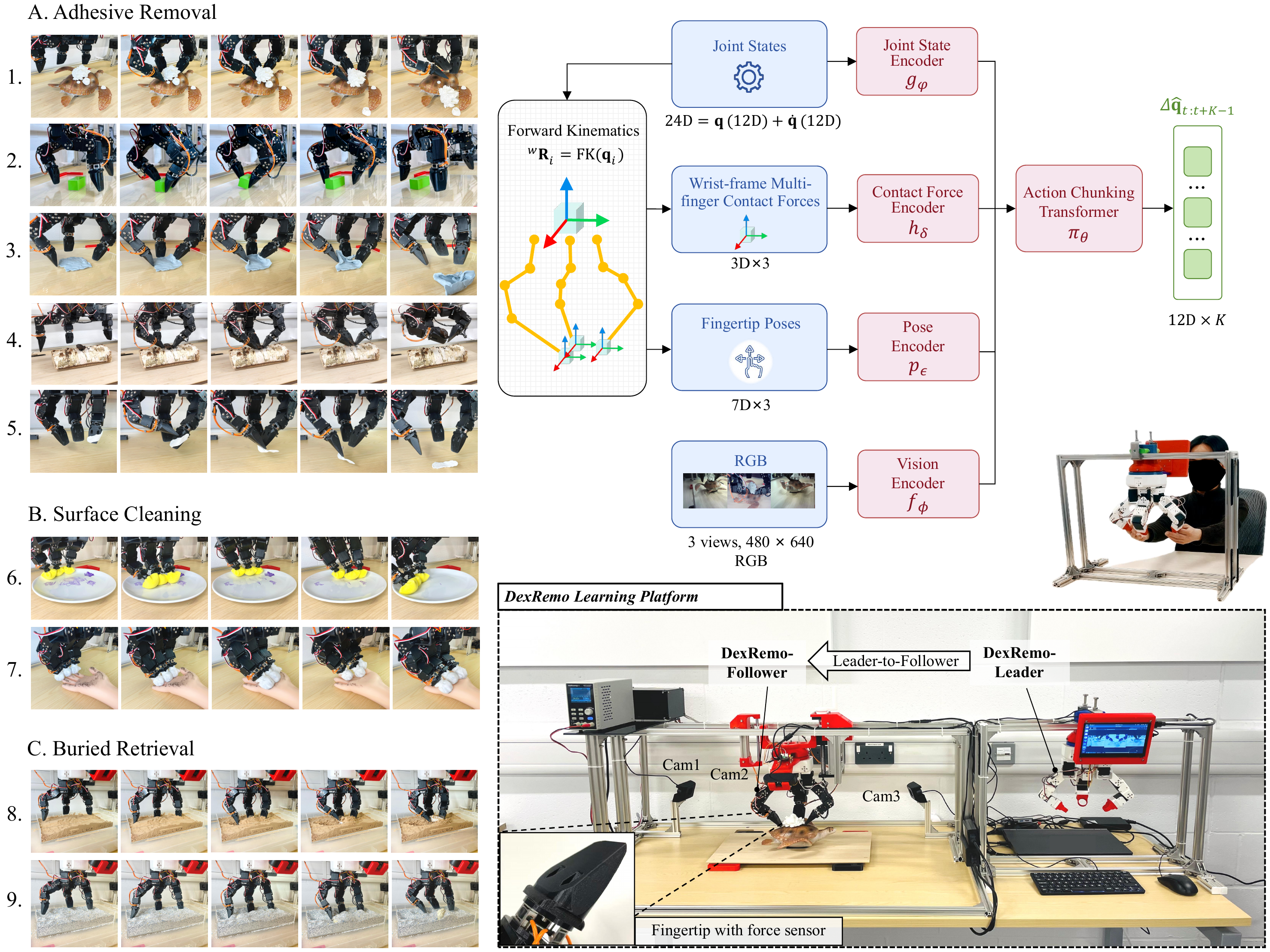}
\caption{
\textbf{Overview of the DexRemo removal manipulation framework.}
\textbf{Left:} Representative sequences from the nine tasks, grouped into the three task families
A (adhesive removal, A1--A5), B (surface cleaning, B1--B2), and C (buried retrieval, C1--C2);
here the uppercase letters denote task families rather than figure panels.
\textbf{Top right:} Multimodal learning pipeline. Joint states, wrist-frame multi-finger contact forces, fingertip poses, and RGB images are encoded and fused to condition the Action Chunking Transformer policy $\pi_\theta$, which predicts short-horizon joint-displacement chunks.
\textbf{Bottom right:} DexRemo learning platform. A bimanual leader--follower teleoperation setup supports demonstration collection and policy replay. The follower robot integrates multiple RGB cameras and custom fingertips with embedded three-axis force sensors.
}
\label{fig_2}
\end{figure*}

\section{Removal Manipulation and DexRemo Platform}
\label{section3}

Removal manipulation spans adhesion, residue removal, self-cleaning, and occluded retrieval. Across these settings, a robot must progressively separate, expose, or clean a target through sustained contact. This section defines the local removal setting and its sensing and control requirements, then presents the DexRemo hand, fingertip force sensing, and wrist-frame force representation used in the experiments.

\subsection{Problem Scope and Design Requirements}
\label{sec:task_abstraction}
\label{sec:structural_properties}

We study \emph{removal manipulation}: contact-rich tasks in which a robot separates, exposes, or cleans a target through physical interaction with its surrounding environment. Unlike grasp-centric manipulation, where the target is often directly accessible, removal manipulation requires overcoming constraints such as adhesion, frictional wedging, deformation, burial, or surface contamination. In this work, we focus on local removal behaviors after the hand is positioned within a task-relevant workspace, rather than on large-workspace end-effector placement or whole-arm mobile manipulation.

Across the suite, success is defined by physical criteria such as complete detachment, sufficient cleaning, or target exposure, while avoiding excessive damage, violent ejection, or loss of contact. The relevant target--environment state includes attachment, occlusion, deformation, frictional constraint, residue distribution, and burial state; the policy does not explicitly estimate this physical state.

The removal tasks considered in this work exhibit four recurring properties. First, progress is \emph{history-dependent}: previous contacts may weaken an adhesive bond, expose an edge, deform surrounding material, spread residue, or uncover part of a buried object. Second, the contact state is \emph{partially observable}: adhesion strength, subsurface constraints, local deformation, and frictional wedging are difficult to infer from RGB observations or joint positions alone. Third, removal involves \emph{multi-finger contact--motion coupling}: different fingers may probe, anchor, scrape, pull, or stabilize the target along different effective motion directions. Fourth, successful behavior requires \emph{continuous force regulation}: insufficient force may fail to initiate separation, whereas excessive or poorly directed force may damage the target, eject the object, or lose contact.

Accordingly, the system combines a force-capable platform and task suite with physical success criteria, a common coordinate frame for multi-finger force measurements across changing finger configurations, and a multimodal policy that replans short action chunks as contact conditions change.

\subsection{Sensing and Policy Implications}
\label{sec:observation_design}

At time $t$, the policy receives a synchronized multimodal observation
\begin{equation}
\mathbf{o}_t =
\left(
\mathbf{I}_t,
\mathbf{q}_t,
\dot{\mathbf{q}}_t,
\mathbf{p}_t,
\mathbf{f}_t
\right),
\label{eq:observation}
\end{equation}
where $\mathbf{I}_t$ denotes multi-view RGB observations, $\mathbf{q}_t$ and $\dot{\mathbf{q}}_t$ denote joint positions and velocities, $\mathbf{p}_t$ denotes wrist-frame fingertip poses, and $\mathbf{f}_t$ denotes stacked fingertip force measurements. The sensing implementation is detailed below, and the policy architecture is presented in Sec.~\ref{section5}.

Force sensing exposes contact events that are difficult to infer from vision and kinematics alone, but the local frames in which these forces are measured rotate with finger configuration. We therefore express fingertip forces in a common wrist-centered frame using forward kinematics, and evaluate wrist-frame, local-frame, and no-force inputs through controlled ablations in Sec.~\ref{section6}.
The four properties in Sec.~\ref{sec:structural_properties} also motivate a temporally adaptive policy. We use ACT~\cite{Zhao2023ACT} as a sequence-prediction backbone,
\begin{equation}
\pi_\theta:
\mathbf{o}_{t-H+1:t}
\mapsto
\Delta \hat{\mathbf{q}}_{t:t+K-1},
\label{eq:policy}
\end{equation}
which maps an observation context of length $H$ to a predicted $K$-step sequence of joint displacements executed in a receding-horizon loop. Throughout, a hat denotes a quantity predicted by the policy. In our implementation, $H=1$, so the policy conditions on the current synchronized multimodal observation.

\begin{figure}[!t]
\centering
\includegraphics[width=1\linewidth]{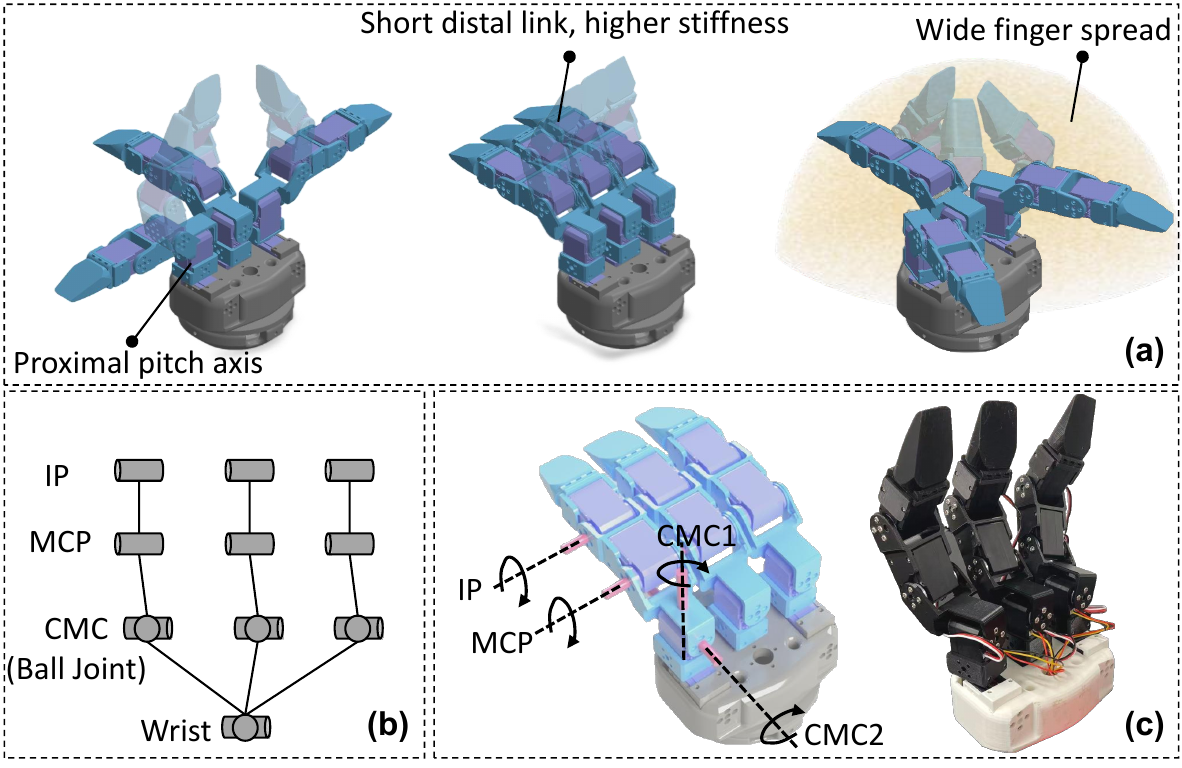}
\caption{
\textbf{DexRemo Hand design.}
(a) Kinematic layout of the three-finger hand.
(b) Finger structure with two flexion joints and a dual-axis base.
(c) Physical prototype with three identical thumb-like fingers for multi-finger manipulation. IP, MCP, and CMC denote the interphalangeal, metacarpophalangeal, and carpometacarpal joints, respectively; CMC1 and CMC2 denote the two axes of the compound CMC joint.
}
\label{fig_4}
\end{figure}

\subsection{Three-Finger Platform Design and Operation}
\label{sec:trifinger}
\label{sec:leader_follower_platform}

DexRemo is a low-cost three-finger robotic hand designed for local removal behaviors such as boundary exploration, edge engagement, scraping, wiping, and progressive detachment. It serves as a reproducible reference platform for studying fingertip force sensing and multimodal learning in removal manipulation.

A three-finger layout provides non-collinear contacts for probing, anchoring, scraping, pulling, and stabilizing while keeping the mechanism simpler and easier to teleoperate than higher-DoF anthropomorphic hands. This morphology is well matched to the partial, asymmetric, and sliding contacts common in peeling, scraping, wiping, and uncovering.

As shown in Fig.~\ref{fig_4}(a)--(b), each finger has four degrees of freedom: a distal interphalangeal (IP) joint, a metacarpophalangeal (MCP) joint, and a two-DoF compound carpometacarpal (CMC) joint. The three fingers are arranged radially to form a hemispherical reachable workspace. Each finger has a total length of $162\,\mathrm{mm}$, with link lengths $[62,50,49]\,\mathrm{mm}$. All joints use Feetech STS3215 servos with closed-loop magnetic encoders, run at $200\,\mathrm{Hz}$, and exhibit command-to-motion latency below $5\,\mathrm{ms}$. The full hand weighs approximately $1.8\,\mathrm{kg}$ and uses high-strength polymer finger structures with aluminum joint housings; the assembled prototype is shown in Fig.~\ref{fig_4}(c).

DexRemo is operated as a leader--follower platform for scalable demonstration collection (Fig.~\ref{fig_2}). The leader device is a passive kinematic replica of the hand, and its joint angles are mapped one-to-one to the follower after per-session zero-offset calibration. The operator coordinates the three fingers using both hands and observes the workspace through external and wrist-mounted camera views, together with a real-time visualization of the follower hand configuration and fingertip force vectors. No haptic feedback is provided. Commands are streamed at $15\,\mathrm{Hz}$ with low end-to-end latency, enabling responsive contact regulation during demonstrations. Demonstration collection, filtering, and training usage are described in Sec.~\ref{section5} and Sec.~\ref{section6}.

\subsection{Modular Fingertips, Force Sensing, and Characterization}
\label{sec:fingertip_sensing}
\label{sec:force_sensor_calibration}

\begin{figure}[!t]
\centering
\includegraphics[width=0.76\linewidth]{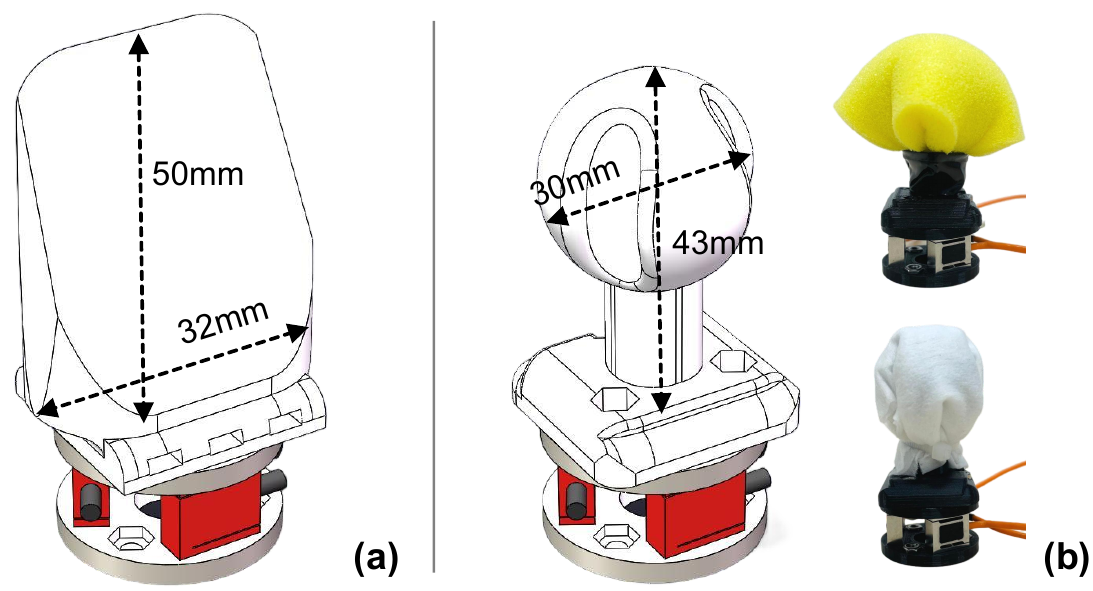}
\caption{
Fingertip tool modules for removal manipulation.
(a) Wedge-shaped rigid fingernail tip integrated with a custom three-axis force sensor.
(b) Soft cleaning tip consisting of a spherical compliant core wrapped with sponge or wet tissue.
Annotated dimensions are shown for reference.
}
\label{fig:tips}
\end{figure}

Each finger uses a modular fingertip unit with interchangeable rigid or compliant shells. Rigid wedge-shaped tips facilitate insertion, scraping, peeling, and prying, while compliant inserts support wiping and cleaning (Fig.~\ref{fig:tips}(a) and (b), respectively). The tool interface is mechanically decoupled from the sensing module, so changing fingertip shells does not require sensor recalibration. Each sensing unit costs approximately \$170, and the force-sensing modules are optional for cost-sensitive deployments.

\begin{figure}[!t]
\centering
\includegraphics[width=1\linewidth]{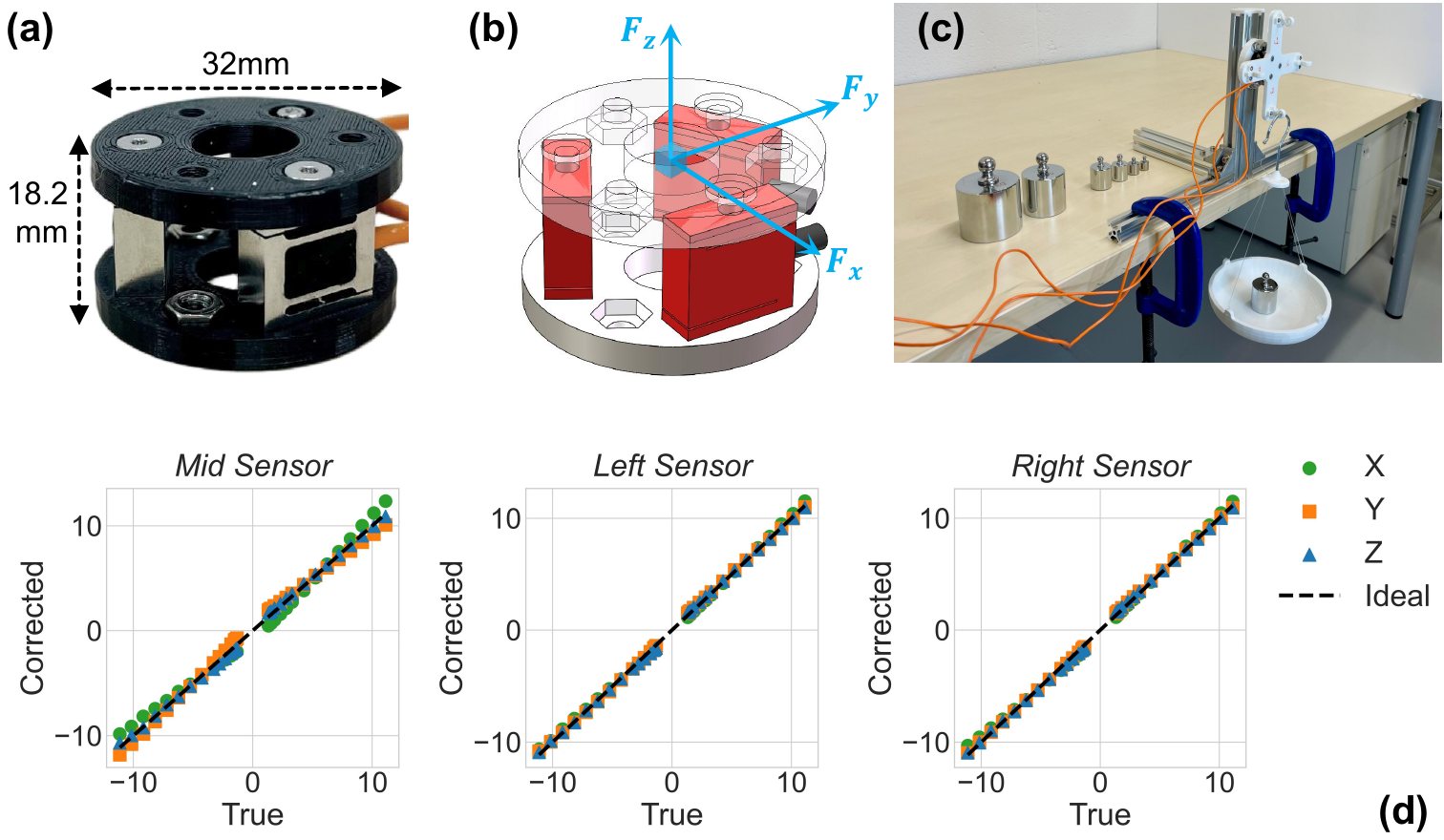}
\caption{
Fingertip three-axis force sensor.
(a) Fabricated sensor with triangular load-cell arrangement.
(b) Exploded view of the rigid frame and measurement axes.
(c) Calibration with suspended masses applying known Cartesian forces.
(d) Calibration for mid/left/right sensors: corrected outputs vs.\ ground-truth forces along $x,y,z$ (dashed: ideal).
}
\label{fig:ft_sensor}
\end{figure}

Each fingertip integrates a compact three-axis force sensing module composed of three industrial miniature S-type strain-gauge load cells (HZC-MS, 0--5\,kg nominal axial range, Chengying Sensor, China). As shown in Fig.~\ref{fig:ft_sensor}, the load cells are arranged in an equilateral triangular configuration within a rigid aluminum frame, allowing reconstruction of normal and tangential contact forces applied to the fingertip after module-level calibration. The manufacturer specifications characterize the individual load cells; after assembly, the complete fingertip module is calibrated and validated separately because mechanical coupling, contact location, mounting preload, and the reconstruction model can affect the final three-axis output.

The sensors measure axial forces $f_1$, $f_2$, and $f_3$ along their local $Z$ directions. Under a rigid contact assumption, the normal component is computed as
\begin{equation}
F_z = f_1 + f_2 + f_3,
\label{eq:fingertip_fz}
\end{equation}
and the planar components are obtained by a calibrated linear mapping,
\begin{equation}
\begin{bmatrix}
F_x \\ F_y
\end{bmatrix}
=
\mathbf{A}
\begin{bmatrix}
f_1 \\ f_2 \\ f_3
\end{bmatrix}
+ \mathbf{b},
\label{eq:fingertip_fxy}
\end{equation}
where $\mathbf{A}\in\mathbb{R}^{2\times3}$ and $\mathbf{b}\in\mathbb{R}^{2}$ are identified from module-level calibration data. Detailed reconstruction assumptions are provided in Appendix~\ref{app:ft_reconstruction}.

The fingertip sensing module provides three-axis translational force measurements while maintaining a compact and mechanically robust form factor. Accordingly, this study focuses on normal and tangential force feedback during removal manipulation. The potential contribution of additional fingertip torque measurements remains an interesting direction for future investigation.

Calibration is performed at the assembled fingertip-module level using externally applied reference forces (Fig.~\ref{fig:ft_sensor}(c)--(d)). Full calibration procedures, reconstruction assumptions, and per-axis error analyses are provided in Appendix~\ref{app:ft_reconstruction}. The learned mappings achieve $R^2 > 0.986$ across all axes (Table~\ref{tab:calib_xyz_full}), providing reliable three-axis contact force estimates for closed-loop manipulation and learning.

Table~\ref{tab:force_sensor_characterization} separates individual load-cell manufacturer specifications, data-acquisition module specifications, and assembled module-level sensing metrics, because the final three-axis force output also depends on the mechanical arrangement, mounting preload, and calibration. The acquisition module provides a 19-bit effective bipolar ADC resolution over a $\pm25\,\mathrm{mV}$ differential input range, while the practical module-level noise floor is characterized by the post-calibration no-load force standard deviation of $0.0034\,\mathrm{N}$ on average and $0.0129\,\mathrm{N}$ at maximum across the tested sensor-axis recordings. 

\begin{figure}[!t]
\centering
\includegraphics[width=1\linewidth]{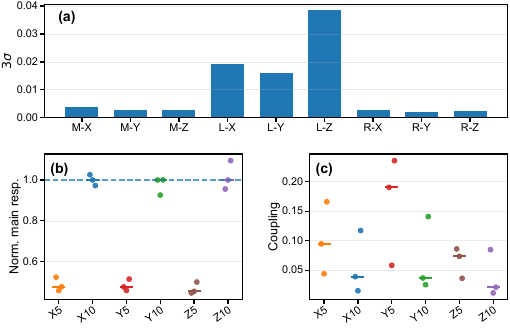}
\caption{
Assembled fingertip force-sensing characterization.
Here, M, L, and R denote the middle, left, and right fingertip modules, respectively.
(a) \(3\sigma\) no-load variation after mean-offset removal.
(b) Static 5~N/10~N main-axis response, normalized by the median 10~N response for each axis; bars denote medians.
(c) Cross-axis coupling, defined as off-axis response norm over absolute main-axis response.
}
\label{fig:sensor_characterization}
\end{figure}

Fig.~\ref{fig:sensor_characterization} further characterizes the measured three-axis force outputs under static no-load and 5~N/10~N loading conditions. After removing the no-load mean offset, the measurements show low output variation, monotonic main-axis response, and quantifiable residual cross-axis coupling.

\begin{table}[t]
\centering
\caption{Fingertip force sensing: manufacturer specifications and
assembled-module characterization.}
\label{tab:force_sensor_characterization}
\footnotesize
\renewcommand{\arraystretch}{1.05}
\setlength{\tabcolsep}{4.5pt}
\begin{tabularx}{\columnwidth}{@{}>{\raggedright\arraybackslash}X c@{}}
\toprule
\textbf{Metric} & \textbf{Value} \\
\midrule
\multicolumn{2}{@{}l}{\textit{Individual load-cell manufacturer specifications}} \\
Rated axial capacity & $5\,\mathrm{kgf}$ ($\approx49\,\mathrm{N}$) \\
Rated output & $2.0 \pm 10\%\,\mathrm{mV/V}$ \\
Zero balance & $\pm 2\%$ F.S. \\
Nonlinearity & $0.15\%$ F.S. \\
Hysteresis & $0.05\%$ F.S. \\
Repeatability & $0.05\%$ F.S. \\
Safe / ultimate overload & $150\%$ / $300\%$ F.S. \\
\midrule
\multicolumn{2}{@{}l}{\textit{Data-acquisition module manufacturer specifications}} \\
ADC effective resolution & $19$ bit, bipolar \\
Differential input range & $\pm25\,\mathrm{mV}$ \\
\midrule
\multicolumn{2}{@{}l}{\textit{Assembled three-axis fingertip module}} \\
Calibration $R^2$ (all reconstructed axes) & $>0.986$ \\
Aggregate force RMSE (N) & $0.349$ \\
No-load force standard deviation (N) & $0.0034$ / $0.0129$ \\
Cross-axis response ratio (\%) & $8.2$ / $23.6$ \\
Max peak normal load $F_z$ in suite (N) & $63.1$ \\
\bottomrule
\end{tabularx}
\vspace{1mm}
\begin{minipage}{0.96\columnwidth}
\footnotesize
\vspace{1mm}
F.S. denotes full scale. The first two blocks list manufacturer
specifications for the individual load cells and data-acquisition
module, respectively. The final block reports measurements obtained
after assembly and calibration of the three-axis fingertip module.
The listed ADC resolution characterizes signal digitization and should
not be interpreted as the minimum resolvable force of the assembled
module. Slash-separated values denote mean / maximum across
sensor-axis recordings. The maximum peak normal load is the largest
per-rollout peak of $F_z = f_1 + f_2 + f_3$ observed anywhere in the
task suite; because this normal load is shared among the three load
cells, the corresponding per-cell axial load is roughly one third of
it and remains well within the rated capacity listed above.
\end{minipage}
\end{table}

\subsection{Wrist-Frame Multi-Finger Force Representation}
\label{sec:wrist_force_representation}

A practical challenge in articulated multi-finger sensing is that fingertip sensor frames rotate continuously with finger motion. Consequently, local-frame force components can change their coordinate meaning even when the underlying physical load is similar, complicating cross-finger comparison and multimodal fusion. To provide consistent directional semantics for learning, we express all fingertip forces in a shared wrist-centered frame.

Using an end-effector or task-centered frame is consistent with classical force control and the Task Frame Formalism, which represent force directions relative to task-relevant coordinates~\cite{vill,bruyninckx1996taskframe,salisbury1982articulated}. In DexRemo, the wrist frame also serves as the common reference for fingertip poses, proprioception, and wrist-mounted visual observations. Let ${}^w\mathbf{R}_{i,t}$ denote the rotation from the $i$-th fingertip sensor frame to the wrist frame at time $t$. The measured fingertip force ${}^i\mathbf{f}_{i,t}$ is transformed as
\begin{equation}
{}^w\mathbf{f}_{i,t}
=
{}^w\mathbf{R}_{i,t}\,{}^i\mathbf{f}_{i,t}.
\label{eq:wrist_force_transform}
\end{equation}
The policy force input is the stacked vector
\begin{equation}
\mathbf{f}_t =
\left[
{}^{w}\mathbf{f}_{1,t};
{}^{w}\mathbf{f}_{2,t};
{}^{w}\mathbf{f}_{3,t}
\right]
\in \mathbb{R}^{9}.
\label{eq:wrist_frame_force_input}
\end{equation}

This transformation is deterministic and does not add sensory information. Its role is to inject a known geometric prior before learning: force components expressed in the wrist frame have consistent directional meaning across fingers and time. Equivalently, for a fixed task-level force ${}^w\mathbf{f}_{i,t}$, the corresponding local-frame input is $({}^w\mathbf{R}_{i,t})^\top\,{}^w\mathbf{f}_{i,t}$; changing the fingertip orientation therefore changes the local components even when the physical force in the wrist frame is unchanged. A policy given local-frame forces and fingertip pose must implicitly learn this pose-dependent inverse rotation in addition to the task-relevant force--action relationship. With only 20 demonstrations per task, this additional entanglement may encourage posture-dependent correlations that do not transfer reliably across finger configurations. This geometric argument motivates the representation choice and is evaluated through the controlled ablations in Sec.~\ref{section6}; it is not presented as a separate causal proof. A detailed derivation is provided in Appendix~\ref{app:wrist_frame_force}; policy ablations are reported in Sec.~\ref{section6}.

\section{Learning Framework}
\label{section5}

We use ACT~\cite{Zhao2023ACT} as a temporal imitation-learning backbone for contact-rich removal manipulation. Given synchronized visual, proprioceptive, fingertip-pose, and force observations, the policy predicts short-horizon joint-space displacement chunks and executes them in a receding-horizon loop.

\subsection{Observation Space and Multimodal Encoding}
\label{sec:obs_preprocess}
\label{sec:multimodal_encoders}

At each time step $t$, the policy uses the synchronized multimodal
observation $\mathbf{o}_t$ defined in Eq.~\eqref{eq:observation}.
Specifically, $\mathbf{I}_t$ denotes synchronized multi-view RGB images,
$\mathbf{q}_t \in \mathbb{R}^{12}$ and
$\dot{\mathbf{q}}_t \in \mathbb{R}^{12}$ denote joint positions and
velocities, respectively, $\mathbf{p}_t$ denotes fingertip poses
expressed in the wrist frame, and $\mathbf{f}_t$ denotes the stacked
three-finger force measurements.
For the full model, the policy uses the stacked wrist-frame force input
$\mathbf{f}_t \in \mathbb{R}^{9}$ defined in
Eq.~\eqref{eq:wrist_frame_force_input}.
For local-frame ablations, the same policy architecture is used, but
$\mathbf{f}_t$ is replaced by the corresponding stacked local-frame
fingertip forces.
For no-force ablations, the force token is removed.

All sensory streams are synchronized to the execution frequency of $15\,\mathrm{Hz}$ before training and deployment. RGB images are represented as tensors of size
$3\times480\times640$ (channel $\times$ height $\times$ width) and normalized using the mean and standard deviation computed from the training demonstrations. Joint positions, joint velocities, fingertip poses, and force measurements are normalized using the mean and standard deviation computed from the training demonstrations of each task. Fingertip poses are represented as wrist-frame Cartesian positions and unit quaternions for the three fingertips, giving $\mathbf{p}_t \in \mathbb{R}^{21}$. 
Forces are clipped to $[-60,60]\,\mathrm{N}$ before normalization to reduce the influence of rare impact spikes.

Each modality is encoded independently into a token representation before fusion by the ACT backbone. The vision stream consists of three RGB camera views. Each view is processed by a ResNet-18 encoder pretrained on ImageNet, and the final spatial feature map is projected to a $d$-dimensional token using a linear layer.
The three camera tokens are then concatenated with the non-visual modality tokens.

The proprioceptive stream concatenates joint positions and velocities,
\begin{equation}
\mathbf{s}_t = [\mathbf{q}_t;\dot{\mathbf{q}}_t] \in \mathbb{R}^{24},
\end{equation}
and maps them to a token through a two-layer MLP. Fingertip poses $\mathbf{p}_t$ are encoded by a separate two-layer MLP, and fingertip forces $\mathbf{f}_t$ are encoded by another two-layer MLP. All non-visual encoders use hidden dimension $256$, ReLU activations, and layer normalization after the final projection. 

Formally, the modality tokens are
\begin{align}
\mathbf{z}^{\mathrm{img}}_t &= f_{\phi}(\mathbf{I}_t), \\
\mathbf{z}^{\mathrm{joint}}_t &= g_{\varphi}([\mathbf{q}_t;\dot{\mathbf{q}}_t]), \\
\mathbf{z}^{\mathrm{pose}}_t &= p_{\epsilon}(\mathbf{p}_t), \\
\mathbf{z}^{\mathrm{force}}_t &= h_{\delta}(\mathbf{f}_t),
\end{align}
where all tokens are projected to the same embedding dimension $d=512$.
For ablation settings without fingertip pose or force input, the corresponding token is removed while all other architectural and training settings are kept unchanged.

\subsection{ACT-Based Chunked Policy}
\label{sec:chunked_policy}

The encoded tokens from an observation context of length $H$ are fused by an ACT transformer. With $H=1$, the mapping of Eq.~\eqref{eq:policy} is instantiated on the current observation as
\begin{equation}
\pi_\theta:
\mathbf{o}_{t}
\mapsto
\Delta \hat{\mathbf{q}}_{t:t+K-1},
\qquad K=100 .
\label{eq:act_policy_mapping}
\end{equation}

The action at each step is a 12-dimensional joint displacement command,
\begin{equation}
\Delta \hat{\mathbf{q}}_{t+j} \in \mathbb{R}^{12}, \quad j=0,\ldots,K-1.
\end{equation}

Following ACT~\cite{Zhao2023ACT}, the model uses a conditional variational structure during training. A latent variable $\mathbf{z}$ is inferred from the demonstration action chunk and conditions the transformer decoder, enabling the policy to represent recurring action patterns in teleoperated demonstrations. At inference time, the prior mean is used for deterministic execution.

The predicted displacement chunk is integrated from the current joint configuration:
\begin{equation}
\hat{\mathbf{q}}_{t+j}
=
\mathbf{q}_t
+
\sum_{\ell=0}^{j-1}
\Delta \hat{\mathbf{q}}_{t+\ell},
\quad j=1,\ldots,K.
\label{eq:chunk_integration}
\end{equation}
During deployment, only the first $k=50$ actions of each predicted $K=100$ step chunk are executed before the policy is queried again with updated observations. This receding-horizon execution allows the policy to revise future actions in response to newly observed visual and force feedback.

\subsection{Demonstration Collection, Training, and Deployment}
\label{sec:demo_collection}
\label{sec:training_objective}
\label{sec:deployment_protocol}

Demonstrations are collected with a leader--follower teleoperation interface. The leader is a passive DexRemo replica with the same joint topology and link geometry. After per-session zero-offset calibration, its encoder-measured joint angles are mapped one-to-one to the follower, which tracks the commanded joint positions using low-level position control.
The demonstrator observes synchronized external and wrist-mounted camera views, together with a real-time hand-state visualization showing the follower hand configurations and three-dimensional fingertip force vectors (Appendix~\ref{app:calibration}, Fig.~\ref{fig:teleop_interface}). These visualizations provide augmented feedback about contact direction and magnitude; no kinesthetic or tactile haptic feedback is provided.
Before data collection, operators complete a short task-specific familiarization phase. We retain only demonstrations that satisfy the task success criterion without visible damage or unsafe contact. We do not assume optimal demonstrations, but use them as successful behavior-cloning examples. For each task, we collect $20$ successful demonstrations under nominal training conditions with variations in object placement and attachment configuration.

Training is performed by supervised behavior cloning from teleoperated demonstrations. Given an observation context $\mathbf{o}_{t-H+1:t}$, the network predicts a $K$-step sequence of joint displacements
\begin{equation}
\Delta \hat{\mathbf{q}}_{t:t+K-1}
=
(\Delta \hat{\mathbf{q}}_{t},\ldots,\Delta \hat{\mathbf{q}}_{t+K-1}),
\end{equation}
which is integrated to obtain the predicted joint trajectory chunk $\hat{\mathbf{q}}_{t+1:t+K}$ using Eq.~\eqref{eq:chunk_integration}. The behavior cloning loss is
\begin{equation}
\mathcal{L}_{\mathrm{BC}}
=
\frac{1}{K}
\sum_{j=1}^{K}
\left\|
\hat{\mathbf{q}}_{t+j}
-
\mathbf{q}_{t+j}
\right\|_1,
\label{eq:bc_loss}
\end{equation}
where $\mathbf{q}_{t+j}$ is the demonstrated joint configuration.

When using the ACT latent variable, we add the standard KL regularization term:
\begin{equation}
\mathcal{L}
=
\mathcal{L}_{\mathrm{BC}}
+
\beta
D_{\mathrm{KL}}
\left(
q_{\psi}
\left(
\mathbf{z}\mid
\mathbf{o}_{t-H+1:t},
\mathbf{q}_{t:t+K}
\right)
\,\|\, 
p(\mathbf{z})
\right),
\label{eq:act_loss}
\end{equation}
where $q_{\psi}$ is the approximate posterior over $\mathbf{z}$ and $\beta=10$.
Although the loss is defined on joint trajectories, the predicted actions are conditioned on wrist-frame fingertip forces together with the other sensory inputs. During deployment, the policy is queried again after the executed action prefix using updated observations. Thus, in DexRemo, force influences subsequent motion through observation-conditioned receding-horizon prediction, rather than through an auxiliary force-prediction loss or explicit force/wrench commands.

Training samples are drawn uniformly from the demonstration trajectories. We train a separate task-specific policy for each task and use the same architecture and hyperparameters across input-configuration ablations unless otherwise stated. The main policy and ablation settings are trained with three random seeds, and evaluation rollouts use the checkpoint saved after $60{,}000$ gradient steps. Table~\ref{tab:policy_hyperparameters} lists the complete policy architecture and training configuration used throughout the experiments.

At inference time, joint-displacement commands are executed at $15\,\mathrm{Hz}$. The policy predicts a $K=100$ step chunk from the current observation and is re-invoked every $k=50$ action steps with updated observations. The first $k$ actions of the current chunk are streamed to the low-level joint-position controller. No task-specific state machine, contact threshold, scripted recovery behavior, or manually designed force controller is used during policy execution.

Generalization is evaluated zero-shot under the held-out object geometries, spatial configurations, colors, and attachment conditions defined in Sec.~\ref{section6}.

\begin{table}[t]
\centering
\caption{Policy architecture and training hyperparameters.}
\label{tab:policy_hyperparameters}
\begin{tabular}{l c}
\toprule
\textbf{Item} & \textbf{Value} \\
\midrule
Policy backbone & ACT~\cite{Zhao2023ACT} \\
Camera views & 3 RGB views \\
Image resolution (height $\times$ width) & $480 \times 640$ \\
Vision encoder & ResNet-18 \\
Vision pretraining & ImageNet \\
Non-visual encoder & Two-layer MLP \\
Non-visual hidden dimension & 256 \\
Non-visual activation & ReLU \\
Non-visual output normalization & LayerNorm \\
VAE latent dimension & 32 \\
VAE encoder layers & 4 \\
Joint input dimension & 24 \\
Fingertip pose dimension & 21 \\
Force input dimension & 9 \\
Token dimension $d$ & 512 \\
Transformer encoder layers & 4 \\
Transformer decoder layers & 1 \\
Attention heads & 8 \\
Feedforward dimension & 3200 \\
Observation context length $H$ & 1 \\
Chunk length $K$ & 100 \\
Executed prefix $k$ & 50 \\
Action dimension & 12 \\
Action execution frequency & $15\,\mathrm{Hz}$ \\
Optimizer & AdamW \\
Learning rate & $1\times10^{-5}$ \\
Weight decay & $1\times10^{-4}$ \\
Batch size & 8 \\
Training steps & 60,000 \\
KL weight $\beta$ & 10 \\
Force clipping range & $[-60,60]\,\mathrm{N}$ \\
Number of demos per task & 20 \\
Random seeds & 3 \\
\bottomrule
\end{tabular}
\end{table}

\section{Experimental Evaluation}
\label{section6}

This section describes the real-robot removal manipulation task suite, evaluation protocol, and results. The suite contains nine tasks in three constraint families: adhesive detachment and self-cleaning (A1--A5), surface mark removal (B1--B2), and occluded retrieval in particulate media (C1--C2). The goal is not a material-characterization benchmark, but a reproducible real-robot evaluation setting with specified preparation procedures, physical success criteria, held-out variations, and consistent policy execution.

\subsection{Experimental Platform and Task Suite}
\label{sec:platform_setup}
\label{sec:task_suite}

The experimental platform (Fig.~\ref{fig_2}, bottom right) consists of DexRemo mounted above a tabletop workspace of size $820\,\mathrm{mm} \times 1156\,\mathrm{mm}$. A linear stage provides vertical approach and retreat motions over $z \in [-205,-115]\,\mathrm{mm}$, where the tabletop is defined as $z=0$ and positive $z$ points upward. This setup focuses on local removal behaviors after the hand is positioned within the task-relevant workspace.

Three RGB cameras provide synchronized observations: two external cameras and one wrist-mounted camera. The policy operates directly on multi-view RGB images and does not require metric-accurate camera calibration. Camera poses, synchronization, and preprocessing details are documented relative to the wrist-centered reference frame used for fingertip poses and wrist-frame forces; additional details are provided in Appendix~\ref{app:calibration}. During evaluation, all policies use the same camera configuration, sensing stack, low-level controller, and receding-horizon execution loop described in Sec.~\ref{section5}.

\begin{figure}[t]
  \centering
  \includegraphics[width=1\linewidth]{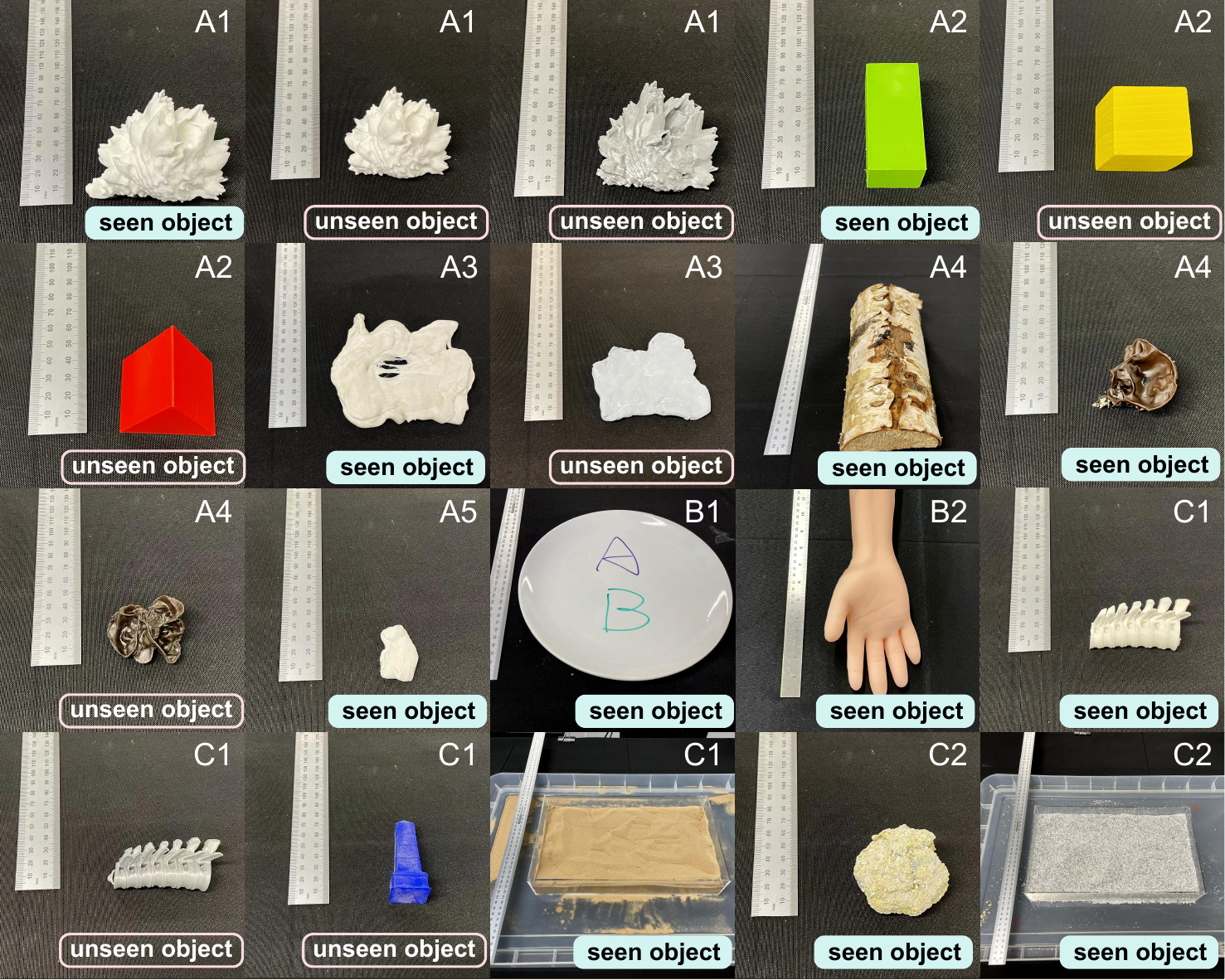}
  \caption{
    Physical objects used in each manipulation task (A1--C2). Each object is photographed alongside a standard ruler to indicate real-world scale.
    Objects are labeled as seen or unseen according to whether they appear in policy training.
    Detailed dimensions and preparation parameters are summarized in Table~\ref{tab:task-preparation} and Appendix~\ref{app:task_details}.
  }
  \label{fig:task_objects}
\end{figure}

The task suite comprises nine removal tasks (Fig.~\ref{fig_2}), with representative physical objects shown in Fig.~\ref{fig:task_objects}. A1--A4 require detaching targets from substrates through adhesive or bonded contacts; A5 evaluates removal of adhered material from the fingertip itself; B1--B2 require progressive surface cleaning; and C1--C2 require exposing or retrieving targets from particulate media. Together, these tasks cover edge acquisition, peeling, scraping, wiping, pulling, digging, and uncovering behaviors.

Table~\ref{tab:task-suite} summarizes the task categories, constraint types, training variations, and held-out tests. Table~\ref{tab:task-preparation} reports the corresponding task objects, materials, and preparation procedures. Additional nominal dimensions, held-out conditions, evaluation conventions, and success criteria are provided in Appendix~\ref{app:task_details}.

\newcolumntype{P}[1]{>{\raggedright\arraybackslash}p{#1}}

\newcommand{\taskthumb}[3]{%
\begin{minipage}[t]{0.315\textwidth}
\centering
\includegraphics[width=\linewidth,height=0.085\textheight,keepaspectratio]{#1}[-0.5mm]
{\scriptsize\textbf{#2} #3}
\end{minipage}
}

\begin{table*}[t]
\centering
\small
\renewcommand{\arraystretch}{1.35}
\setlength{\tabcolsep}{5pt}
\caption{Task set summary with training-time variations and held-out test conditions. Representative task images from the second camera view are shown in the rightmost column.}
\label{tab:task-suite}
\begin{tabular}{m{0.7cm} P{3.2cm} P{3.4cm} P{3.6cm} P{3.8cm} m{2.1cm}}
\toprule
\textbf{ID} & \textbf{Task} & \textbf{Constraint type} & \textbf{Train variations} & \textbf{Held-out tests} & \textbf{Cam2 view} \\
\midrule
\multicolumn{6}{l}{A. Adhesive detachment and self-cleaning} \\
\midrule
A1 & Barnacle removal from turtle shell &
hot-melt adhesive bonding &
barnacle pose &
unseen barnacle size/pose/color &
\raisebox{-0.5\height}{\includegraphics[width=2cm]{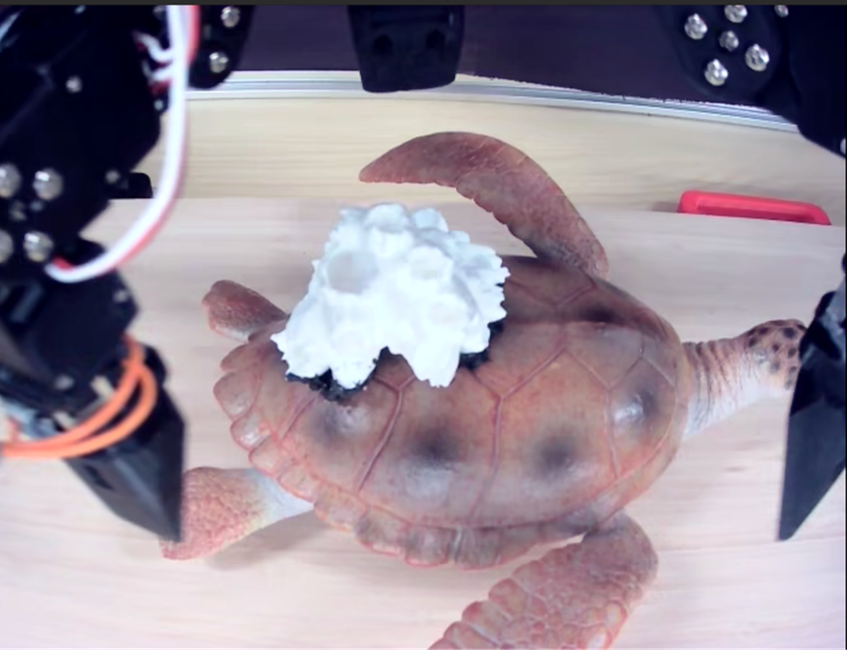}} \\
A2 & 3D-printed part removal from the build plate &
nano-adhesive bonding &
3D-printed part pose &
unseen geometries/color; unseen pose &
\raisebox{-0.5\height}{\includegraphics[width=2cm]{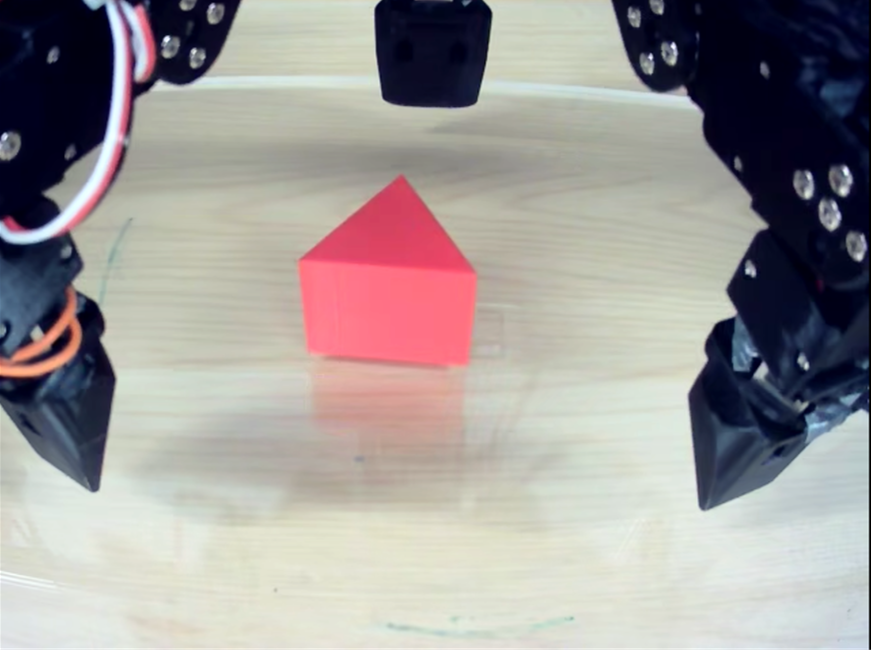}} \\
A3 & Blu-Tack removal from acrylic plate &
Blu-Tack adhesion &
Blu-Tack shape/pose &
unseen Blu-Tack shape/color/placement &
\raisebox{-0.5\height}{\includegraphics[width=2cm]{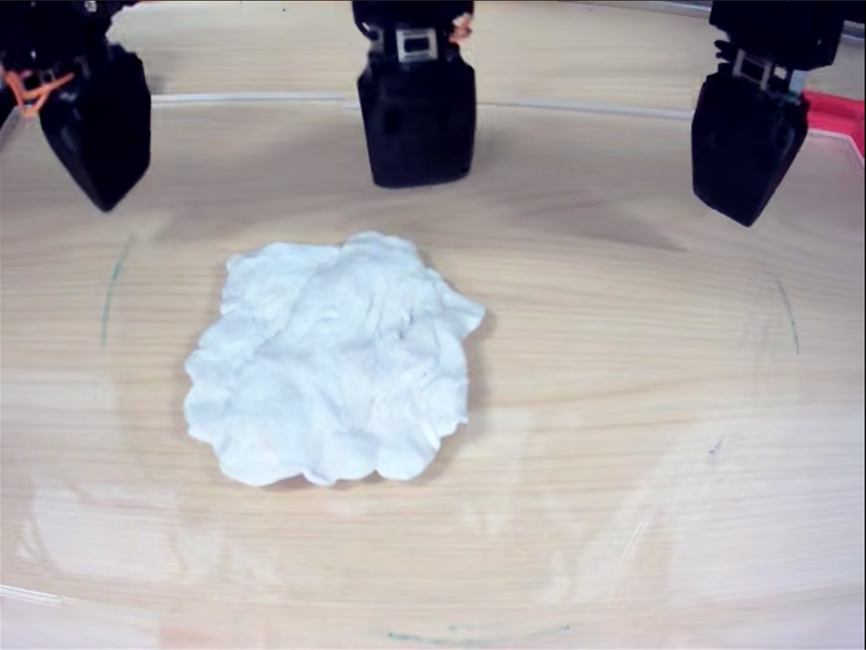}} \\
A4 & Wood-ear extraction from tree log &
hot-melt adhesive bonding &
wood-ear pose &
unseen wood-ear pose/size &
\raisebox{-0.5\height}{\includegraphics[width=2cm]{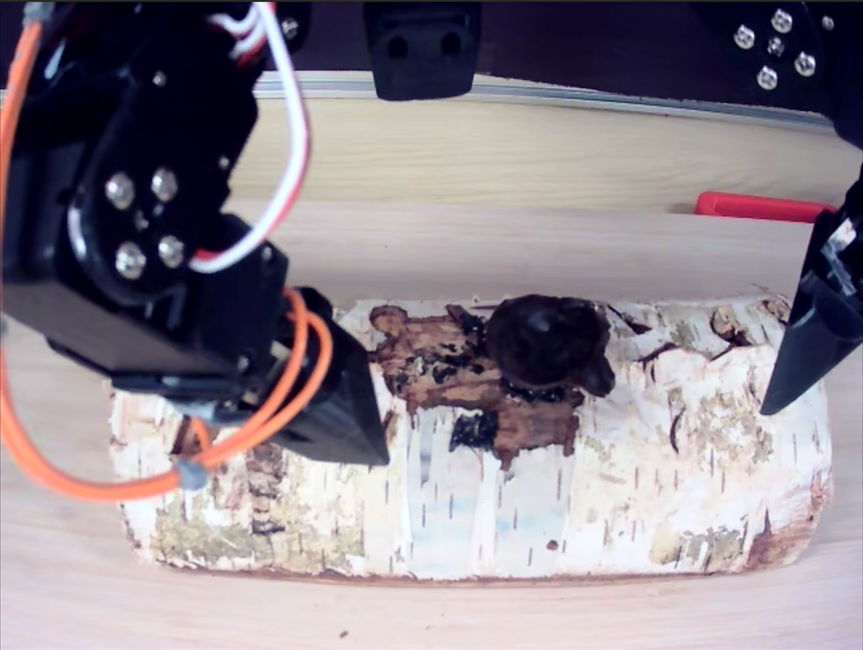}} \\
A5 & Fingertip self-cleaning &
Blu-Tack adhesion &
adhesive residue shape/pose &
unseen adhesive residue shape/pose/color &
\raisebox{-0.5\height}{\includegraphics[width=2cm]{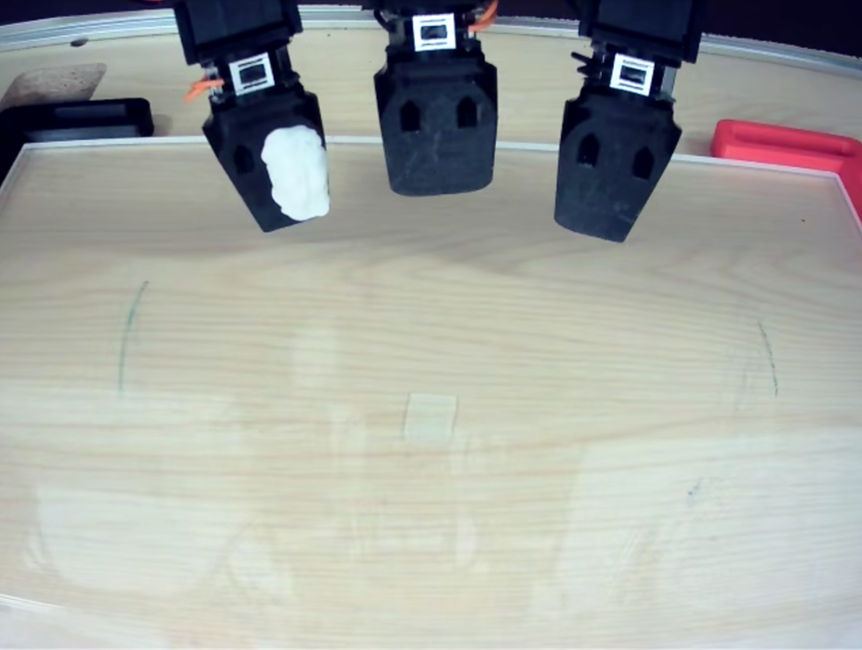}} \\
\midrule
\multicolumn{6}{l}{B. Surface mark removal} \\
\midrule
B1 & Sponge wiping marker stains from the ceramic plate &
pigment surface adhesion &
stain location/geometry &
unseen stain location/geometry/color &
\raisebox{-0.5\height}{\includegraphics[width=2cm]{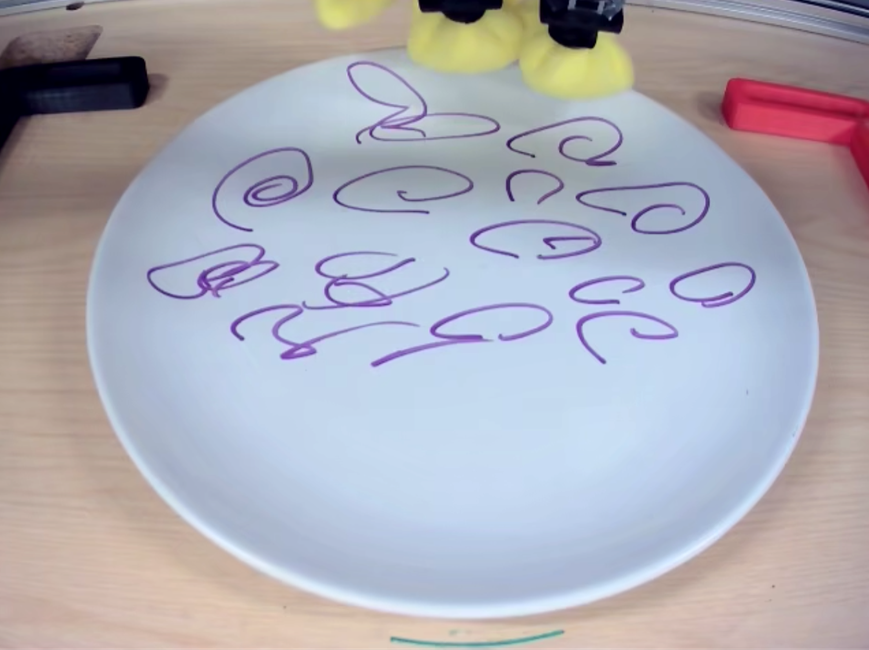}} \\
B2 & Cotton wiping ink marks on silicone skin &
pigment surface adhesion &
stain location/geometry &
unseen stain location/geometry/color &
\raisebox{-0.5\height}{\includegraphics[width=2cm]{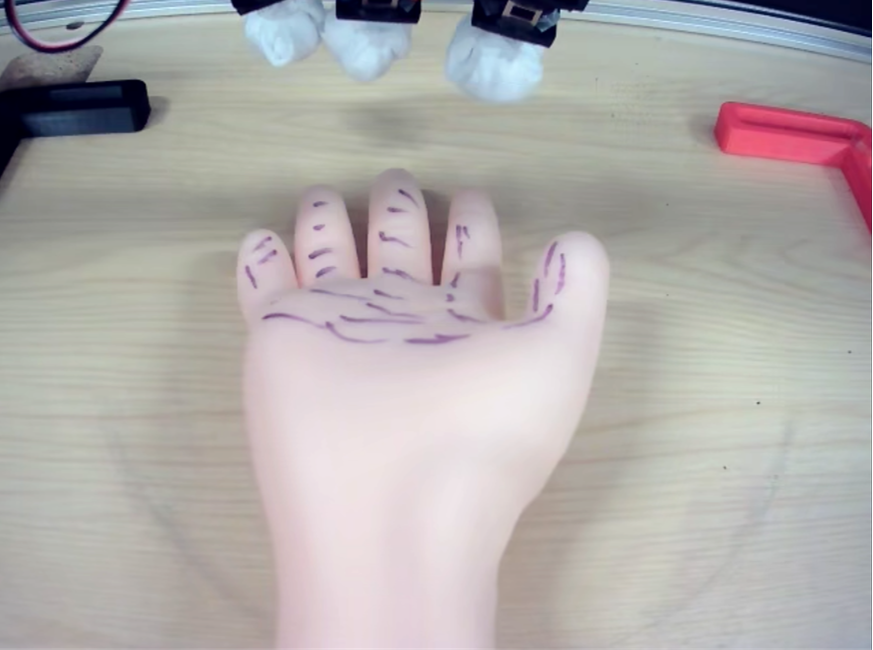}} \\
\midrule
\multicolumn{6}{l}{C. Occluded object retrieval} \\
\midrule
C1 & Fossil excavation in sand &
inter-granular forces in sand &
fossil pose &
unseen pose; unseen fossils &
\raisebox{-0.5\height}{\includegraphics[width=2cm]{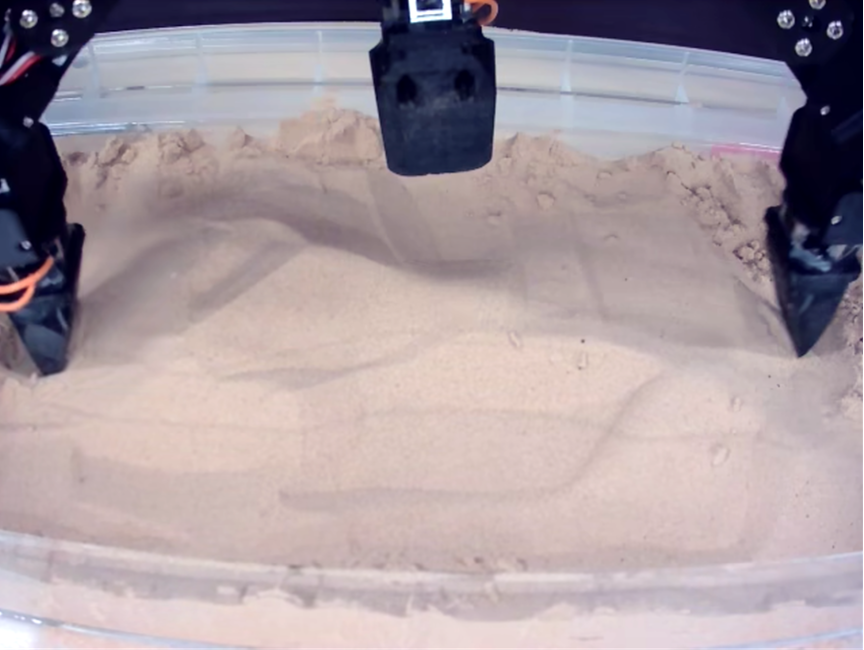}} \\
C2 & Cat litter clump removal &
inter-granular forces in cat litter &
cat litter clump pose/size &
unseen cat litter clump color/pose &
\raisebox{-0.5\height}{\includegraphics[width=2cm]{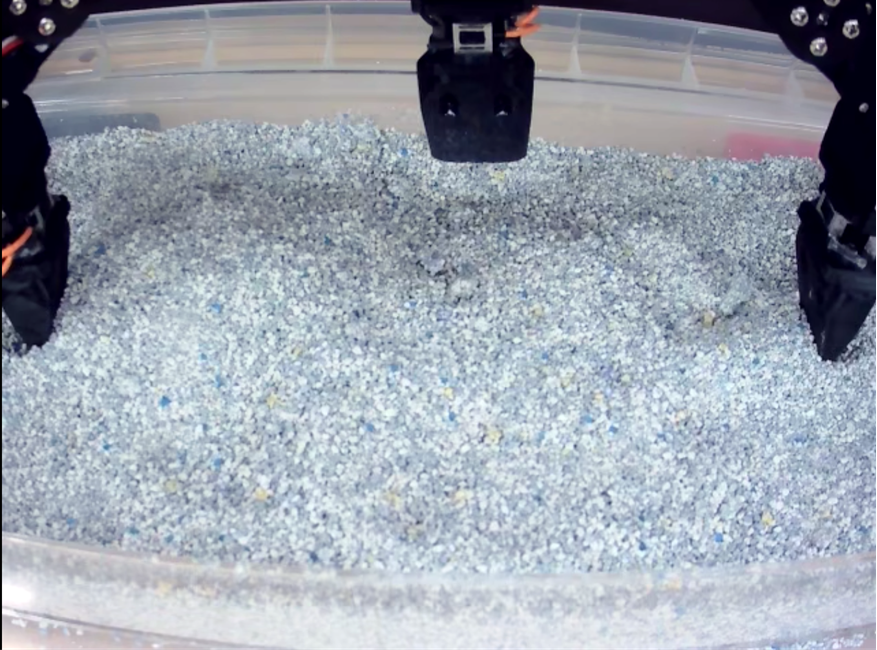}} \\
\bottomrule
\end{tabular}
\end{table*}

\begin{table*}[t]
\centering
\caption{Task objects, nominal specifications, materials, and preparation procedures.}
\label{tab:task-preparation}
\scriptsize
\renewcommand{\arraystretch}{1.12}
\setlength{\tabcolsep}{4pt}
\begin{tabularx}{\textwidth}{c Y Y Y}
\toprule
\textbf{ID} &
\textbf{Object / nominal specification} &
\textbf{Material / medium} &
\textbf{Preparation protocol} \\
\midrule
A1 &
Barnacle model, $50{\times}40{\times}40\,\mathrm{mm}$ &
PLA/resin target on turtle-shell substrate; hot-melt adhesive &
Apply $0.35\,\mathrm{g}$ adhesive and wait $60\,\mathrm{s}$ before trial. \\
A2 &
3D-printed part, $65{\times}30{\times}30\,\mathrm{mm}$; adhesive pad, $15{\times}15\,\mathrm{mm}$ &
PLA part on build plate; nano-adhesive film &
Attach part to cleaned plate and press for $10\,\mathrm{s}$. \\
A3 &
Blu-Tack patch, $150{\times}100{\times}1\,\mathrm{mm}$ ($\approx 20\,\mathrm{g}$) &
Blu-Tack on acrylic plate &
Knead to target shape and press with $5\,\mathrm{N}$ for $5\,\mathrm{s}$. \\
A4 &
Wood-ear model, $50{\times}40\,\mathrm{mm}$ &
Soft target on birch log; hot-melt adhesive &
Apply adhesive at contact root and wait $60\,\mathrm{s}$ before trial. \\
A5 &
Blu-Tack residue, $20{\times}20{\times}5\,\mathrm{mm}$ &
Blu-Tack on fingertip shell &
Attach residue to fingertip and randomize its shape and pose. \\
B1 &
Marker stain, $120{\times}120\,\mathrm{mm}$ target region &
Purple marker on ceramic plate; sponge tip &
Draw stain in predefined region and wet sponge with $2\,\mathrm{ml}$ water. \\
B2 &
Ink stain, $120{\times}120\,\mathrm{mm}$ target region &
Ink on silicone skin; cotton tip &
Draw ink mark and wait $30\,\mathrm{s}$ before trial. \\
C1 &
Fossil model, $80{\times}40{\times}30\,\mathrm{mm}$ &
Target buried in beach sand &
Bury at $15\,\mathrm{mm}$ nominal depth and flatten the surface. \\
C2 &
Cat-litter clump, $70{\times}15{\times}20\,\mathrm{mm}$ &
Clump embedded in cat litter &
Embed the clump and level the surrounding litter. \\
\bottomrule
\end{tabularx}
\vspace{0.5mm}
\end{table*}

\subsection{Task Protocol and Evaluation Procedure}
\label{sec:task_protocol}
\label{sec:eval_protocol}

Each rollout starts from a task-specific reset state prepared according to Table~\ref{tab:task-preparation}. The hand is placed within the local task workspace, the linear stage moves to the task-specific approach height, and the learned policy controls the hand joints through the receding-horizon interface described in Sec.~\ref{section5}. All rollouts terminate when the task success condition is met or when the time limit is reached. No scripted recovery behavior, task-specific contact threshold, or manual intervention is used during evaluation.

Demonstration collection follows Sec.~\ref{section5}. For each scenario, we collect $N_{\text{demo}}=20$ successful teleoperated demonstrations under nominal training conditions. Training variations include object pose, contact location, shape, and attachment configuration, depending on the scenario. Held-out tests use unseen object geometries, colors, poses, or attachment conditions as summarized in Table~\ref{tab:task-suite}. All evaluated pose variations are local perturbations within the hand's reachable workspace and should not be interpreted as large-workspace end-effector, whole-arm, or mobile pose generalization.

We organize evaluation into two tiers. The first tier consists of six core benchmark tasks, A1--A4, B1, and C1, which span the three principal task families and carry the complete six-configuration, three-seed comparison with matched rollout budgets; unless noted otherwise, each core task uses $30$ rollouts per input configuration and random seed. The second tier consists of three extension scenarios, A5, B2, and C2, which broaden the suite to fingertip self-cleaning, compliant-surface cleaning, and cluttered particulate-media retrieval. These are evaluated only under the full $P+W$ configuration with $10$ held-out rollouts per task, because their resets are more costly and less repeatable at scale, or because they involve constraint mechanisms already covered by the core tier. They are therefore reported as additional task coverage rather than matched comparisons, and are excluded from core-task averages and seed-level confidence intervals.

Success rates are reported with 95\% confidence intervals computed across the three seed-level success rates. Comparisons between input configurations are restricted to the six core benchmark tasks, where all input configurations are evaluated with matched rollout counts.

\subsection{Ablation Settings, Evaluation Metrics, and Runtime Environment}
\label{sec:ablations}

To isolate the contributions of fingertip pose, force availability, and force coordinate representation, we train policies under multiple input configurations while keeping the demonstrations, policy architecture, training schedule, and evaluation protocol unchanged. All six settings include multi-view RGB observations and joint proprioception, held fixed as the common observation context. The matched ablations therefore isolate fingertip pose, force availability, and force coordinate representation; the role of vision itself is probed separately in Sec.~\ref{sec:baseline_comparison}. We compare six configurations built from three optional inputs: $P$, fingertip poses in the wrist frame, represented by Cartesian position and unit quaternion for each fingertip; $W$, wrist-frame fingertip forces; and $R$, fingertip-local force vectors. The configurations are the baseline $\emptyset$, which removes both pose and force, together with $R$, $P$, $W$, $R+P$, and the full $P+W$. They test whether gains arise from force sensing itself, fingertip-pose conditioning, or expressing multi-finger forces in a shared coordinate frame.

The primary metric is binary task success under the task-specific physical criteria defined in Appendix~\ref{app:task_details}. Beyond success rate, the following subsections report termination and completion times, completion efficiency, force-impulse-based effort measures, policy inference latency, and failure-mode statistics.

Training is performed on a workstation equipped with an Intel i7-14700 CPU and an NVIDIA RTX 4070 GPU (12\,GB). Policy inference runs on a separate desktop with an AMD R7 5800 CPU and an NVIDIA RTX 2060 GPU (6\,GB). The sensing and action-execution loop operates in real time at $15\,\mathrm{Hz}$ without specialized hardware. The average policy inference latency is $23.5 \pm 4.1\,\mathrm{ms}$ per query.

\subsection{Platform Validation and Multimodal Force-Alignment Results}
\label{sec:platform_validation}
\label{sec:success_modality_ablation}

We first characterize the platform's force capability and operational robustness under the loads induced by the suite.
The hand reaches a peak pullout strength of $62.5\,\mathrm{N}$ and a power density of $3.064$ under the protocol in Appendix~\ref{app:pullout_protocol}, while the largest peak normal fingertip load observed anywhere in the suite is $63.1\,\mathrm{N}$ (Table~\ref{tab:force_sensor_characterization}). Because the normal component is the sum of the three load-cell readings, $F_z = f_1 + f_2 + f_3$, a $63.1\,\mathrm{N}$ normal load corresponds to approximately $21\,\mathrm{N}$ per cell and therefore remains well within the $5\,\mathrm{kgf}$ ($\approx49\,\mathrm{N}$) rated capacity and the safe overload limit of each load cell listed in Table~\ref{tab:force_sensor_characterization}. This peak also exceeds the $[-60, 60]\,\mathrm{N}$ range to which the policy force input is clipped (Table~\ref{tab:policy_hyperparameters}); only isolated peak samples in the most force-demanding cleaning scenario reach this bound, and the same clipping is applied identically to every input configuration, so it does not affect the reported comparisons.
The $90$th-percentile per-rollout peak normal load is $22.2$--$59.0\,\mathrm{N}$ across the seven adhesive-detachment and surface-cleaning scenarios (A1--A5, B1--B2), in which $20$--$100\%$ of rollouts exceed $20\,\mathrm{N}$, and $11.4\,\mathrm{N}$ and $6.6\,\mathrm{N}$ for the two particulate-media scenarios (C1--C2). Tens-of-newton fingertip loads are therefore routine in the adhesive and cleaning families, whereas retrieval in particulate media operates in a much lower force regime.
Module-level sensing characterization is reported in Sec.~\ref{sec:force_sensor_calibration}.
Across the matched ablations, the extension scenarios, and the supplementary diagnostics, the platform completed $3{,}420$ real-robot evaluation rollouts, in addition to the teleoperated demonstrations collected for training, over more than $100$~h of logged operation with $23$ logged hardware interruptions.

We evaluate the six input configurations defined in Sec.~\ref{sec:ablations}.

For task \(j\), input configuration \(m\), and seed \(s\), the seed-level success rate is
\begin{equation}
S_{j,m,s}
=
\frac{1}{N}
\sum_{r=1}^{N} y_{j,m,s,r},
\qquad N=30,
\label{eq:seed_success_rate}
\end{equation}
where \(y_{j,m,s,r}\in\{0,1\}\) indicates whether rollout \(r\) succeeds. We report
\begin{equation}
\bar{S}_{j,m}
=
\frac{1}{N_{\mathrm{seed}}}
\sum_{s=1}^{N_{\mathrm{seed}}} S_{j,m,s},
\qquad N_{\mathrm{seed}}=3,
\label{eq:mean_success_rate}
\end{equation}
with a \(95\%\) confidence interval
\begin{equation}
\bar{S}_{j,m}
\pm
t_{0.975,N_{\mathrm{seed}}-1}
\frac{\mathrm{std}_{s}(S_{j,m,s})}
{\sqrt{N_{\mathrm{seed}}}} .
\label{eq:success_ci}
\end{equation}
These intervals summarize variability across independently trained policies. The 30 rollouts per seed reduce rollout-level uncertainty for each fixed checkpoint, whereas the three training seeds provide a limited basis for model-level statistical inference. We therefore use the intervals to communicate seed-level uncertainty and consistent empirical trends, rather than as formal hypothesis tests on individual pairwise differences. At the task level, however, the ordering is consistent: $P+W$ attains a higher seed-averaged success rate than every other evaluated configuration on all six core tasks. In particular, $P+W$ exceeds its local-frame counterpart $R+P$ on all six tasks, and $W$ exceeds $R$ on all six; each of these two orderings is significant under a two-sided sign test ($p=0.031$).

\begin{figure*}[!t]
\centering
\includegraphics[width=1\linewidth]{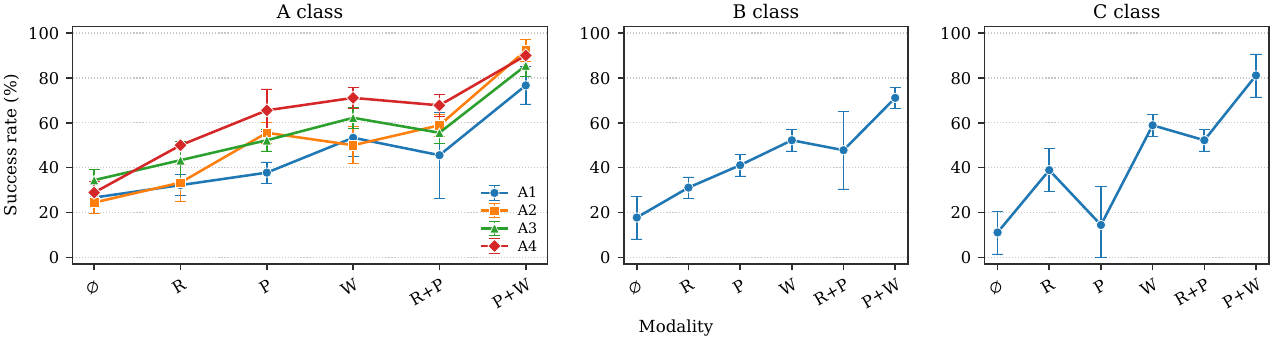}
\caption{
Task-wise success rates across input configurations on the six core benchmark tasks.
All six matched configurations use multi-view RGB observations and joint proprioception; additional inputs include local-frame fingertip forces \(R\), fingertip poses \(P\), and wrist-frame fingertip forces \(W\).
Points show mean success rates over three random seeds, with 30 real-robot rollouts per seed; error bars denote 95\% confidence intervals computed across seed-level success rates.
The full \(P+W\) configuration achieves the strongest overall performance across A1--A4, B1, and C1.
}
\label{fig:success_rate_modality}
\end{figure*}

Fig.~\ref{fig:success_rate_modality} shows success rates across the six core tasks and input configurations. Combining fingertip pose with wrist-frame forces $(P+W)$ yields the strongest and most consistent overall performance among the evaluated configurations. As a descriptive summary over all core-task rollouts, the success rates are $24\%$ ($\emptyset$), $38\%$ ($R$), $44\%$ ($P$), $58\%$ ($W$), $55\%$ ($R+P$), and $83\%$ ($P+W$). Local-frame force does raise mean success over the corresponding no-force configuration, both without fingertip pose ($\emptyset$ to $R$) and with it ($P$ to $R+P$); what the wrist frame adds is a larger and uniform gain, with $W$ exceeding $R$ and $P+W$ exceeding $R+P$ on all six core tasks.
The largest gain from adding wrist-frame force is observed in fossil excavation (C1), where the target is occluded and the measured forces carry contact-state information that the common visual and proprioceptive inputs do not. The gap between local-frame and wrist-frame forces follows the representation-learning argument of Sec.~\ref{sec:wrist_force_representation}: because local components are posture-dependent, a policy given local-frame force and fingertip pose must learn the pose-dependent coordinate transformation together with the task-relevant force--action relationship. This offers a plausible explanation for why local-frame force yields a smaller gain than the same measurements expressed in the wrist frame, and we read it as consistent with the controlled ablations rather than as a separately established causal effect.

\begin{figure}[!t]
\centering
\includegraphics[width=0.8\linewidth]{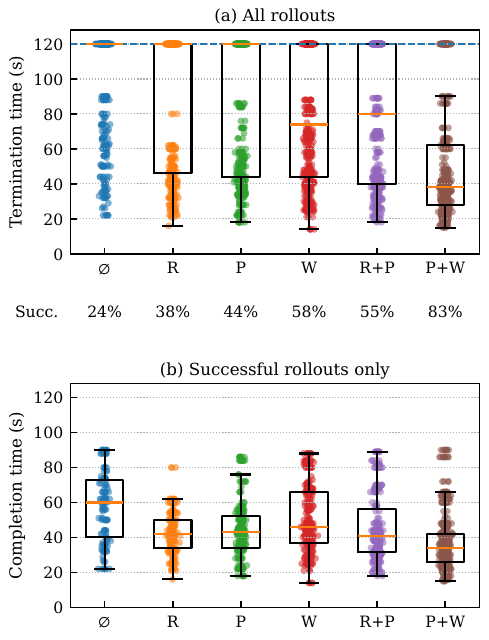}
\caption{
Completion and termination time across input configurations over all core tasks.
(a) Termination time over all rollouts, where failed trials are assigned the 120~s timeout.
The success rate of each input configuration is reported below the x-axis.
(b) Completion time over successful rollouts only.
Boxes show the interquartile range, horizontal lines denote medians, whiskers show the non-outlier range, and dots denote individual real-robot rollouts.
The full \(P+W\) configuration achieves the highest overall success rate and substantially reduces timeout-dominated failures, while maintaining competitive successful-rollout completion time.
}
\label{fig:completion_time_all}
\end{figure}

We further compare execution time across input configurations, using the timing metrics formally defined in Sec.~\ref{sec:task_specific_tradeoff}, Eqs.~\eqref{eq:capped_time}--\eqref{eq:completion_efficiency}. For all-rollout termination time, failed trials are assigned the 120~s timeout, so this metric jointly reflects reliability and execution efficiency. We also report completion time over successful rollouts only to separate speed from failure frequency. As shown in Fig.~\ref{fig:completion_time_all}, the full \(P+W\) configuration achieves the highest overall success rate and the lowest timeout-affected termination distribution. Although successful-only completion times overlap across several input configurations, \(P+W\) maintains a low median completion time while substantially reducing timeout failures. This suggests that wrist-frame forces combined with fingertip pose improve practical execution rather than merely increasing success on a small subset of easy trials.

\subsection{Task-Specific Efficiency--Force Trade-offs}
\label{sec:task_specific_tradeoff}

Success rate alone does not distinguish efficient contact from excessive force use, nor reveal whether force is regulated in a task-appropriate way, so we analyze completion efficiency and force impulse by task class. Impulse is interpreted differently across classes: in A-class adhesive removal and C-class retrieval, high impulse often reflects jamming, ineffective contact, or unnecessary disturbance, whereas in B-class cleaning, sustained impulse can indicate stable pressing and wiping over spatially extended stains.

For each task class \(c\) and input configuration \(m\), we compute the timeout-capped mean termination time,
\begin{equation}
\bar{T}^{\mathrm{cap}}_{c,m}
=
\frac{1}{N_{c,m}}
\sum_{r=1}^{N_{c,m}} T^{\mathrm{cap}}_r ,
\label{eq:capped_time}
\end{equation}
where
\begin{equation}
T^{\mathrm{cap}}_r
=
\begin{cases}
T_r, & s_r=1,\\
T_{\max}, & s_r=0,
\end{cases}
\qquad
T_{\max}=120~\mathrm{s}.
\label{eq:capped_time_rollout}
\end{equation}
The completion-efficiency score is
\begin{equation}
E_{c,m}
=
1 -
\frac{\bar{T}^{\mathrm{cap}}_{c,m}}{T_{\max}} .
\label{eq:completion_efficiency}
\end{equation}
Thus, larger \(E_{c,m}\) indicates fewer timeout failures and shorter termination times.

For task \(j\) under input configuration \(m\), we first define the
mean force-magnitude impulse across evaluation rollouts as
\begin{equation}
I_{j,m}
=
\frac{1}{N_{j,m}}
\sum_{r=1}^{N_{j,m}}
\int_{0}^{T_r}
\sum_{i=1}^{3}
\left\|{}^{w}\mathbf{f}_{i,r}(t)\right\|_{2}
\,\mathrm{d}t ,
\label{eq:mean_force_impulse}
\end{equation}
where \(T_r\) is the recorded duration of rollout \(r\). Unlike the
timeout-capped \(T_r^{\mathrm{cap}}\) used for completion efficiency,
the integral covers only the interval with available force
measurements, and is evaluated numerically from the
\(15~\mathrm{Hz}\) samples. Thus \(I_{j,m}\), in \(\mathrm{N\,s}\), is
the mean cumulative three-finger force-magnitude impulse for task
\(j\) and configuration \(m\).

To compare force use across tasks with different force scales, we
normalize the mean force-magnitude impulse within each task,
\begin{equation}
\tilde{I}_{j,m}
=
\frac{I_{j,m}}
{\max_{m'} I_{j,m'}} ,
\label{eq:task_normalized_impulse}
\end{equation}
and average it within each task class,
\begin{equation}
\bar{I}_{c,m}
=
\frac{1}{|\mathcal{J}_{c}|}
\sum_{j \in \mathcal{J}_{c}}
\tilde{I}_{j,m},
\label{eq:class_normalized_impulse}
\end{equation}
where \(\mathcal{J}_{c}\) denotes the set of tasks belonging to task
class \(c\).

Because force impulse has different physical meanings across task
classes, we define the task-specific force score as
\begin{equation}
Q_{c,m}
=
\begin{cases}
1-\bar{I}_{c,m}, & c \in \{\mathrm{A},\mathrm{C}\},\\
\bar{I}_{c,m},   & c=\mathrm{B}.
\end{cases}
\label{eq:task_specific_force_score}
\end{equation}
For A- and C-class tasks, a higher \(Q_{c,m}\) corresponds to lower
normalized cumulative contact effort and is consistent with less
prolonged or ineffective contact.
For B-class surface-cleaning tasks, a higher \(Q_{c,m}\) corresponds to
greater normalized cumulative contact and is interpreted as stronger
sustained contact during wiping.

\begin{figure*}[!t]
\centering
\includegraphics[width=0.85\linewidth]{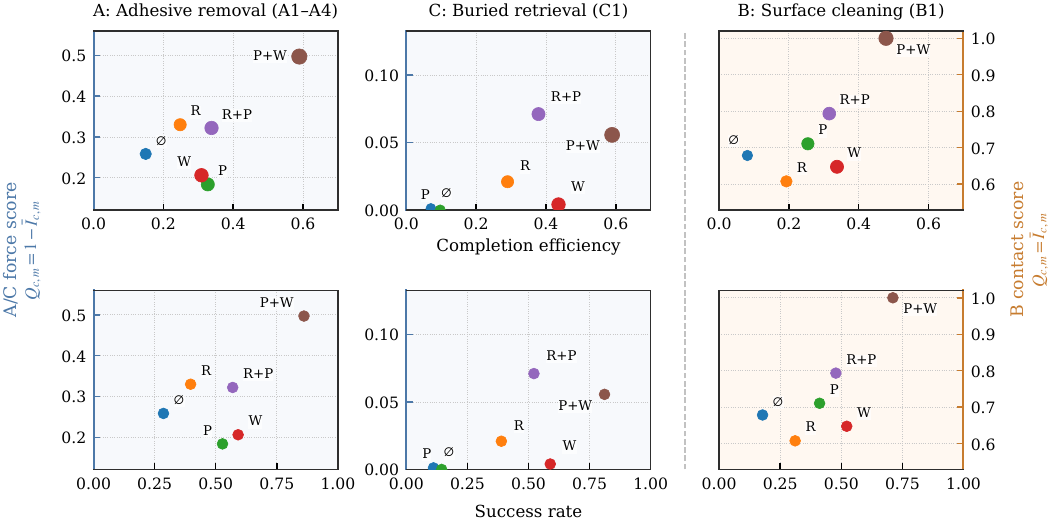}
\caption{
Task-specific efficiency--force/contact trade-off across input
configurations.
Columns correspond to the three core task groups: adhesive removal
(A1--A4), buried retrieval (C1), and surface cleaning (B1).
Each point summarizes one input configuration over the same three
training seeds and matched rollout budgets used in
Fig.~\ref{fig:success_rate_modality}, with rollouts pooled across
seeds before the class-level scores are computed; the point markers
therefore carry no seed-level dispersion.
The top row relates completion efficiency,
Eq.~\eqref{eq:completion_efficiency}, to the class-specific
force/contact score; the bottom row relates success rate to the same
score.
The vertical axis is the task-specific score \(Q_{c,m}\) of
Eq.~\eqref{eq:task_specific_force_score}: for A- and C-class tasks
\(Q_{c,m}=1-\bar{I}_{c,m}\), so a higher score indicates lower
cumulative contact effort, whereas for B-class cleaning
\(Q_{c,m}=\bar{I}_{c,m}\), so a higher score indicates stronger
sustained contact.
Vertical scores should therefore be compared only within a task class.
The full (\(P+W\)) policy shows the most favorable class-specific
trade-offs.
}
\label{fig:tradeoff_grid}
\end{figure*}

Fig.~\ref{fig:tradeoff_grid} shows the resulting trade-offs. In A-class tasks, the full \(P+W\) policy combines high completion
efficiency and reliability with a high A/C force score
(i.e., low normalized cumulative force impulse), suggesting more
efficient edge engagement rather than simply larger force use. In B-class cleaning, \(P+W\) achieves the strongest sustained-contact score, consistent with continuous pressing and wiping over extended stains. In C-class retrieval, force scores are more tightly clustered, indicating that impulse is less discriminative in low-force exploration. Overall, these results suggest that wrist-frame forces and fingertip pose improve task-relevant force regulation rather than uniformly increasing or decreasing contact force.

\subsection{Behavior, Failure Modes, and Extension Scenarios}
\label{sec:failure_modes}

\begin{figure*}[!t]
\centering
\includegraphics[width=1\linewidth]{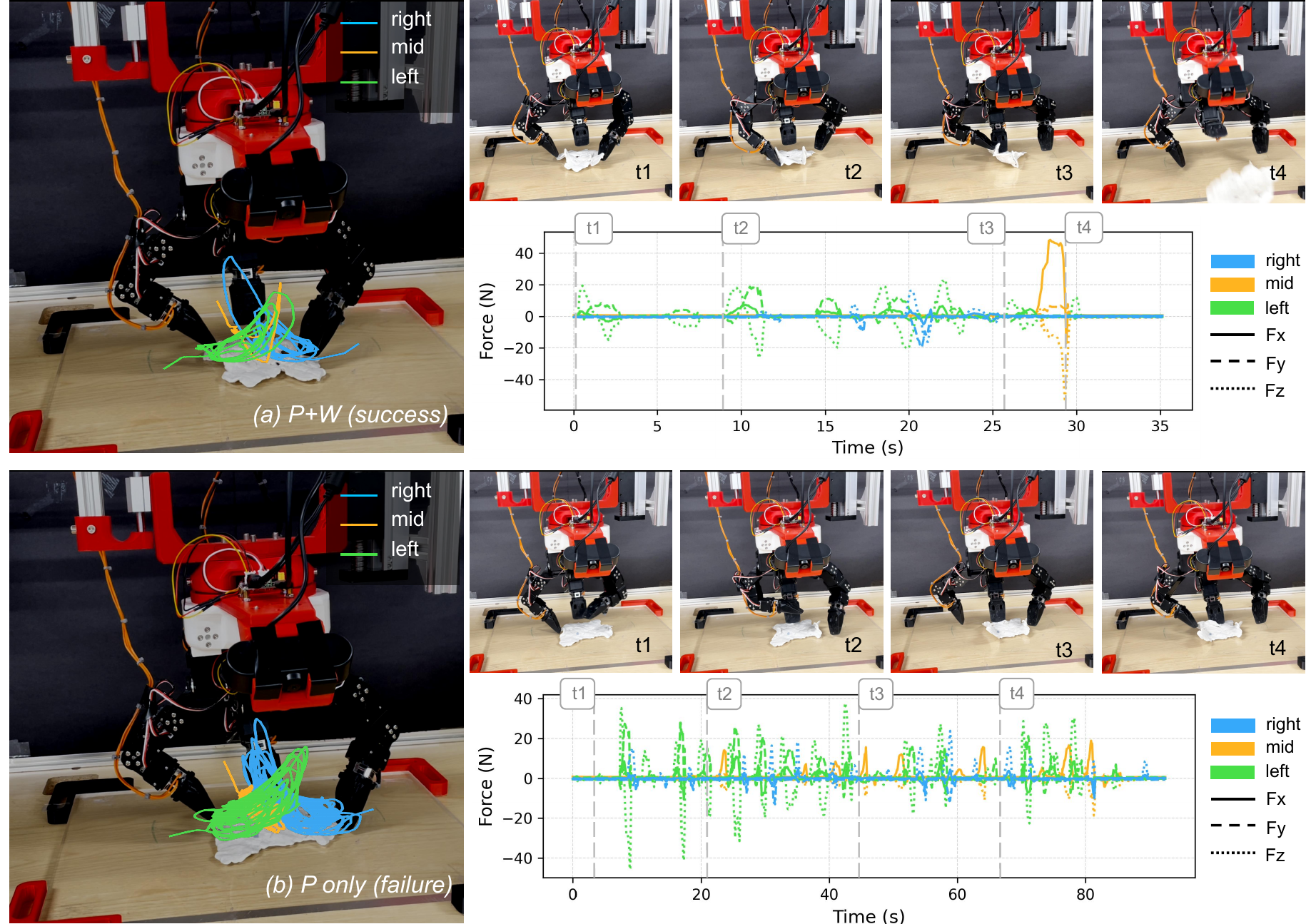}
\caption{Blu-Tack removal (A3): force feedback enables observation-driven switching in an ACT-based policy.
(a) P+W (fingertip pose + wrist-frame forces) and (b) P (fingertip pose only).
Left: fingertip trajectories for right/middle/left fingers. Right: wrist-frame force components over time (color indicates finger; line style indicates $F_x$, $F_y$, and $F_z$).
Vertical dashed lines mark the timestamps of snapshots t1--t4.
With P+W, changes in force patterns during side-finger boundary interaction provide an observable cue for switching to a distinct action chunk dominated by the middle finger, which ramps up force and achieves detachment.
With P only, the policy exhibits prolonged oscillatory contact and fails to satisfy the success criterion within the time limit.}
\label{fig:blu_tack_case}
\end{figure*}

Fig.~\ref{fig:blu_tack_case} shows a representative Blu-Tack removal rollout comparing the full policy $(P+W)$ with a pose-only policy $(P)$. With wrist-frame forces, changes in force patterns during side-finger boundary interaction provide an observable cue for switching to a middle-finger-dominant prying action. Without force feedback, the policy repeats oscillatory contacts and fails to reach a decisive detachment behavior within the time limit.

To quantify how policies fail, we assign each unsuccessful rollout one mutually exclusive dominant failure mode from rollout videos and logs. \emph{Incomplete removal}: partial progress that does not satisfy the physical success criterion, such as residual attachment, remaining contamination, or incomplete retrieval. \emph{Failed engagement}: no effective task-relevant contact is established, such as missing an edge or the contaminated region, or failing to localize the buried target. \emph{Loss of contact}: an initially useful contact later slips away from the task-relevant region. \emph{Excessive force}: overly large or poorly directed forces, including object ejection, visible damage, or unrecoverable jamming. \emph{Insufficient progress}: stable interaction that is too slow to meet the criterion within the time limit.

\begin{figure*}[!t]
\centering
\includegraphics[width=1\linewidth]{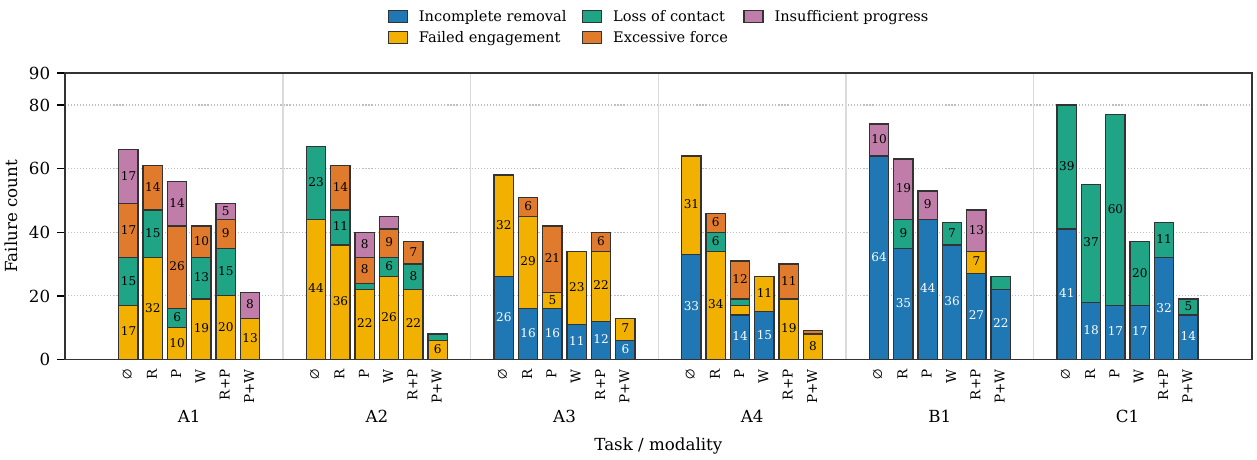}
\caption{
Failure-mode distribution across tasks and input configurations.
Stacked bars show the number of failed rollouts assigned to each dominant failure mode for the six core ablation tasks A1--A4, B1, and C1.
Counts are pooled over the three training seeds for each task--input configuration and are not normalized by total rollouts.
As each input configuration uses the same evaluation budget within a task, lower total bar height indicates fewer failures.
}
\label{fig:failure_type}
\end{figure*}

Fig.~\ref{fig:failure_type} shows not only how often policies fail, but how they fail. Adhesive-removal failures are dominated by residual attachment, ineffective edge engagement, and unstable release under poorly directed force, whereas B1 and C1 fail more often through loss of task-relevant contact or insufficient progress, reflecting the need for sustained surface contact in cleaning and broader spatial exploration in excavation. Across all six core tasks, $(P+W)$ yields the lowest total number of failures of any evaluated input configuration, and the remaining failures are mostly near-misses under strict physical criteria rather than complete behavioral collapse.

We additionally evaluate the three extension scenarios, A5, B2, and C2, under the full $P+W$ configuration with $10$ held-out rollouts each. Because alternative input configurations were not run on these scenarios, their results are reported separately and are excluded from the task-averaged ablation statistics, the comparisons among input configurations, and the seed-level confidence intervals of the six core tasks.

\begin{table*}[t]
\centering
\caption{Extension-scenario evaluation under held-out generalization. A5, B2, and C2 are evaluated only under the full $P+W$ configuration, each with a single trained policy and $10$ held-out rollouts. They are reported as extension evaluations and are not included in the matched input-configuration ablation averages or seed-level intervals. Success rates are reported with $95\%$ binomial confidence intervals. Completion time is averaged over successful rollouts.}
\label{tab:task_success}
\small
\renewcommand{\arraystretch}{1.15}
\setlength{\tabcolsep}{4pt}
\begin{tabularx}{\textwidth}{c X X c c}
\toprule
\textbf{ID} & \textbf{Scenario} & \textbf{Held-out factors} & \textbf{Success rate} & \textbf{Completion time} \\
\midrule
A5 & Fingertip self-cleaning & Residue shape, pose, and color &
$8/10$; $80.0\%$ [$49.0,94.3$] &
$68.5 \pm 20.4\,\mathrm{s}$ \\
B2 & Ink cleaning on silicone skin & Stain location, geometry, and color &
$9/10$; $90.0\%$ [$59.6,98.2$] &
$57.9 \pm 15.3\,\mathrm{s}$ \\
C2 & Cat litter clump exposure & Clump pose and color-matching condition &
$7/10$; $70.0\%$ [$39.7,89.2$] &
$86.7 \pm 19.6\,\mathrm{s}$ \\
\bottomrule
\end{tabularx}
\end{table*}

As shown in Table~\ref{tab:task_success}, the full multimodal policy also transfers to the three extension scenarios, but the remaining failures reveal limits in coverage and long-horizon exploration. In fingertip self-cleaning (A5), residue may remain on fingertip regions that the other fingers cannot reach due to hand morphology and joint limits, making full surface coverage difficult. In silicone-skin cleaning (B2), residual ink is mainly left in narrow grooves or creases, where contact access is constrained and the remaining stain can be partially occluded in the camera views. In cat litter clump exposure (C2), failures typically occur when the policy explores too locally and does not cover enough area before timeout.

\subsection{Supplementary Diagnostics}
\label{sec:baseline_comparison}

We compare ACT with a Diffusion Policy baseline~\cite{dp} using the same full multimodal input---fingertip pose, wrist-frame force (\(P+W\)), multi-view vision, and proprioception---and the same demonstrations. The comparison retains the fixed three-task subset used in the original submission: A2, A3, and B1, covering rigid-part detachment, deformable adhesive removal, and sustained surface cleaning. Each backbone is evaluated from a single seed with \(N_{\mathrm{test}}=30\) real-robot rollouts per task; the ACT column reuses the seed-1 \(P+W\) policy of Sec.~\ref{sec:platform_validation}, so both backbones share identical inputs, demonstrations, and rollout budgets. Because only one checkpoint is evaluated per backbone and the subset does not cover particulate-media retrieval, this experiment is reported as a matched-input, matched-rollout diagnostic rather than a comprehensive policy-family comparison.

The Diffusion Policy conditions on two consecutive observations, predicts 16 actions, and executes four before replanning; it uses an ImageNet-1K-pretrained ResNet-18 on full-resolution \(480\times640\) frames without cropping, matching the ACT vision front end. The DDIM scheduler uses 100 training and five inference denoising steps. Optimization uses a learning rate of \(1\times10^{-4}\), weight decay of \(1\times10^{-6}\), cosine scheduling with 500 warm-up steps, batch size four, and \(100{,}000\) training steps, against \(60{,}000\) steps at batch size eight for ACT. Each backbone otherwise runs at the chunk configuration standard for its family, so the prediction horizons differ: ACT predicts \(K=100\) and executes \(k=50\) from a single observation. The complete configuration is included in the released code.

Table~\ref{tab:baseline_comparison} shows higher observed success rates for the evaluated ACT checkpoint on all three tasks in the fixed diagnostic subset. However, the result may remain sensitive to backbone-specific optimization and tuning, and we do not interpret it as general evidence that ACT is superior to Diffusion Policy. Because the two backbones also differ in prediction horizon, the gap should not be read as isolating the decoder mechanism from temporal abstraction. The main analysis remains the controlled within-ACT comparison of fingertip inputs and force-coordinate representations.

The matched ablations cannot isolate the role of vision, because RGB is a common input in all six configurations. We therefore retrained \(P+W\) with the RGB streams replaced by black frames on two pre-selected core tasks: A3, where the patch varies in shape and placement, and B1, where the stain may lie anywhere within the target region. Both fall to \(0/30\) from their same-seed \(P+W\) references. Without visual localization the hand never reaches the contact states force regulates; vision is necessary here, though the floor precludes quantifying its contribution. This single-seed probe is excluded from all matched comparisons and aggregate statistics.

\begin{table}[t]
\centering
\caption{
Diagnostic comparison of ACT and Diffusion Policy under the same full multimodal \(P+W\) input configuration.
Each entry reports successful rollouts out of 30 real-robot trials.
}
\label{tab:baseline_comparison}
\small
\renewcommand{\arraystretch}{1.15}
\setlength{\tabcolsep}{5pt}
\begin{tabular}{lcc}
\toprule
\textbf{Task} & \textbf{ACT} & \textbf{Diffusion Policy} \\
\midrule
A2 & $28/30$ & $6/30$ \\
A3 & $25/30$ & $9/30$ \\
B1 & $21/30$ & $7/30$ \\
\bottomrule
\end{tabular}
\end{table}

\section{Discussion and Conclusion}
\label{section8}

\subsection{Limitations and Future Work}

DexRemo currently studies local tabletop removal after the hand has been positioned within a task-relevant workspace. The setup uses a fixed wrist pose and a linear stage for vertical motion, and therefore does not address large-workspace end-effector placement, whole-arm manipulation, or mobile manipulation in unstructured scenes. These local separation skills complement global positioning from a robot arm, gantry, or mobile platform, and extending beyond the present setting will require integration with such systems, together with active perception, workspace-level planning, and exploration strategies for locating removable targets and feasible contact regions.

The proposed suite is designed for reproducible real-robot policy evaluation rather than fully specified material characterization. Although we specify preparation procedures, reset rules, held-out variations, and physical success criteria, real removal tasks still contain variability in adhesion, contamination, deformation, and particulate media. Future work should further standardize material preparation, attachment strength, surface state, and cross-laboratory protocols.

The matched ablations do not isolate the standalone contribution of vision, since multi-view RGB and joint proprioception are common inputs; the single-seed probe in Sec.~\ref{sec:baseline_comparison} is directional only. A systematic vision-dependent versus vision-free comparison across task families is left for future work.

As a deterministic geometric alignment, the wrist-frame representation provides an inductive bias for multimodal learning with 20 demonstrations per task without adding sensory information or replacing richer contact sensing. Future work should compare alternative frames, learned alignment mechanisms, dense tactile images, contact-state embeddings, and fingertip wrench sensing. Finally, our policies are trained offline from teleoperated demonstrations without haptic feedback, online adaptation, or uncertainty-aware exploration. The real-time visualization of fingertip force vectors assists the demonstrator but does not replace kinesthetic or tactile feedback and does not guarantee demonstration optimality; accordingly, we use the retained demonstrations as successful behavior-cloning examples rather than optimal trajectories. Future systems could incorporate haptic feedback, improved demonstration filtering, confidence-aware execution, online fine-tuning, and broader transfer across tasks, tools, hands, arms, and sensing modalities.

\subsection{Conclusion}
\label{conclusion}

We presented DexRemo, a force-sensing hand platform and nine-scenario real-robot task suite for contact-rich removal manipulation. The suite uses six core benchmark tasks for controlled fingertip-input and force-frame ablations, together with three extension scenarios for additional held-out evaluation.
The controlled ablations indicate that force representation is an important practical design choice for multi-finger imitation learning in the evaluated 20-demonstration-per-task setting. A shared wrist-frame force representation, especially when combined with fingertip pose, provides the strongest and most consistent stability and generalization compared with the evaluated local-frame and no-force configurations.
The results further show that force-sensitive removal behaviors can be learned from successful teleoperated demonstrations without task-specific state machines or hand-designed force heuristics.

\bibliographystyle{IEEEtran}
\bibliography{refs}

\appendices

\section{Pullout Strength Measurement Protocol}
\label{app:pullout_protocol}

The metrics used in this protocol are summarized in Table~\ref{tab:pullout}.
We measure pullout strength via a quasi-static pull test: the hand is held at a fixed
configuration with constant joint commands while an external force gauge applies a
monotonic pull along the separation direction.
Pullout strength is the peak force immediately before failure (slip or release).
Unless noted, measurements are taken at the middle phalanx for consistency with
prior work; fingertip forces are also recorded but not used for cross-system comparison.
Power density is
\[
\text{Power Density}=\frac{F_{\text{peak}}\cdot \text{DOF}}{A},
\]
where $F_{\text{peak}}$ is pullout strength, DOF is the number of actuated DoFs, and
$A$ is the projected hand area (cm$^2$)\cite{shaw2023leap}.

\section{Three-Axis Fingertip Force Reconstruction}
\label{app:ft_reconstruction}

Each fingertip integrates three single-axis load cells in a rigid triangular layout, as shown in Fig.~\ref{fig:ft_sensor}. The cells measure axial forces $f_i$ along fixed local directions $\mathbf{n}_i$. Assuming a rigid frame and small deformations, each measurement is approximated as a linear projection of the resultant force,
\(
f_i \approx \mathbf{n}_i^\top \mathbf{F} + \epsilon_i,
\)
which motivates the linear reconstruction model in Sec.~\ref{sec:fingertip_sensing}. Under ideal symmetry, the normal component is obtained by summation and the in-plane components by a linear mapping. In practice, this mapping is identified from module-level calibration with externally applied reference forces, so that the fitted parameters capture geometry, cross-axis coupling, mounting effects, and bias. Table~\ref{tab:calib_xyz_full} reports the calibration fit and error metrics.

\begin{table}[htbp]
\centering
\caption{Calibration of in-house force sensors (X/Y/Z): mean squared error (MSE),
$R^2$, and linear fit $y=ax+b$.}
\renewcommand{\arraystretch}{1.18}
\setlength{\tabcolsep}{6pt}
\begin{tabular}{llccc}
\toprule
\textbf{Sensor} & \textbf{Axis} & \textbf{MSE ($\mathrm{N}^2$)} & $\boldsymbol{R^2}$ & \textbf{Fit} \\
\midrule
Mid & X & 0.4924 & 0.9865 & $y = 1.193x - 0.406$ \\
Mid & Y & 0.2788 & 0.9924 & $y = 0.924x + 0.894$ \\
Mid & Z & 0.0656 & 0.9982 & $y = 0.960x + 0.484$ \\
Left & X & 0.0511 & 0.9986 & $y = 1.171x - 0.373$ \\
Left & Y & 0.0220 & 0.9994 & $y = 1.077x + 0.296$ \\
Left & Z & 0.0314 & 0.9991 & $y = 0.862x - 0.003$ \\
Right & X & 0.0978 & 0.9973 & $y = 1.111x - 0.299$ \\
Right & Y & 0.0199 & 0.9995 & $y = 1.000x - 0.251$ \\
Right & Z & 0.0371 & 0.9990 & $y = 0.892x - 0.357$ \\
\bottomrule
\end{tabular}
\label{tab:calib_xyz_full}
\end{table}

\section{Wrist-Frame Force Representation}
\label{app:wrist_frame_force}

To keep force and motion components geometrically consistent~\cite{bruyninckx1996taskframe}, we express all
fingertip forces in a shared wrist-centered frame $\{w\}$. Let $\{i\}$ be the local frame
of fingertip sensor $i$. At time $t$, forward kinematics gives
${}^{w}\mathbf{R}_{i,t}\in SO(3)$. For a measured force ${}^{i}\mathbf{f}_{i,t}$,
\begin{equation}
{}^{w}\mathbf{f}_{i,t} = {}^{w}\mathbf{R}_{i,t}\,{}^{i}\mathbf{f}_{i,t}.
\label{eq:app_wrist_force}
\end{equation}
This change of basis yields force directions consistent across finger configurations and
time, enabling stable multi-finger fusion and force--motion association in removal tasks.
If the sensor frame $\{s_i\}$ differs from $\{i\}$ by a fixed mounting rotation
${}^{i}\mathbf{R}_{s_i}$, then
\begin{equation}
{}^{w}\mathbf{f}_{i,t} =
{}^{w}\mathbf{R}_{i,t}\,{}^{i}\mathbf{R}_{s_i}\,{}^{s_i}\mathbf{f}_{i,t},
\end{equation}
where ${}^{i}\mathbf{R}_{s_i}$ is set by mounting geometry and calibration.

\section{Camera Setup and Synchronization}
\label{app:calibration}

We use three RGB cameras (two external, one wrist-mounted) at $15\,\mathrm{Hz}$ and
$480{\times}640$ (height $\times$ width), synchronized by a shared timestamp source with inter-camera offsets
below $50\,\mathrm{ms}$. Poses are specified in a wrist-centered frame $\{w\}$ at the
mounting flange (tabletop $z{=}0$, $+z$ upward), used for viewpoint reporting and
wrist-frame forces (Sec.~\ref{sec:wrist_force_representation}); the policy itself does not rely on explicit
geometric projections. The nominal poses are:
left external $[{150},{440},{180}]$, $[{0,5,15}]$;
right external $[{150},{-450},{180}]$, $[{0,-5,-15}]$;
wrist-mounted $[{-200},{0},{-30}]$, $[{45,0,0}]$
(position in mm, orientation in degrees).
Unless otherwise stated, no task-specific segmentation, masking, or geometric
rectification is applied. The linear stage provides vertical motion over
$z\in[-205,-115]\,\mathrm{mm}$.

\begin{figure*}[!t]
\centering
\includegraphics[width=0.9\linewidth]{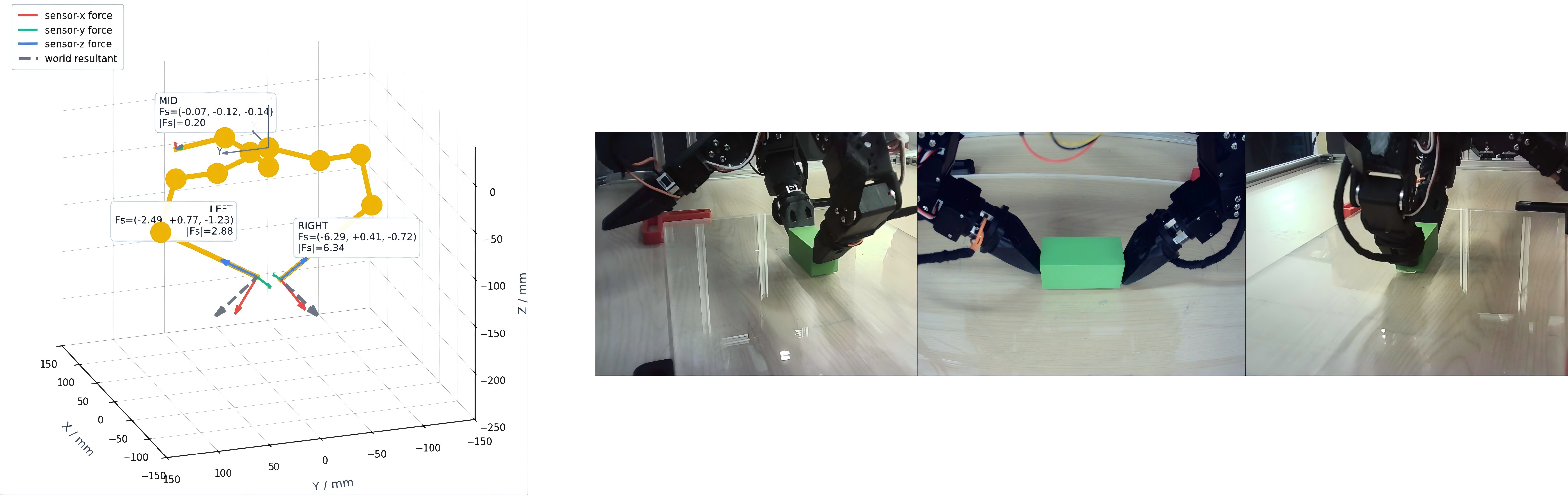}
\caption{Teleoperation interface used during demonstration collection. The demonstrator observes synchronized camera views together with a real-time visualization of the follower hand configuration and fingertip force vectors, which provides augmented visual feedback about contact direction and magnitude.}
\label{fig:teleop_interface}
\end{figure*}

Fig.~\ref{fig:teleop_interface} shows the teleoperation interface, including synchronized camera views and a real-time visualization of the follower hand configuration and fingertip force vectors. No kinesthetic or tactile haptic feedback is provided.

\begin{table*}
\centering
\caption{Additional task details and evaluation criteria. Object identities and constraint types for each task are given in Table~\ref{tab:task-suite} and Table~\ref{tab:task-preparation}.}
\label{tab:task_details_physical}
\scriptsize
\renewcommand{\arraystretch}{1.06}
\setlength{\tabcolsep}{3pt}
\begin{tabular}{@{}L{0.7cm} L{5.3cm} L{5.7cm} L{5.2cm}@{}}
\toprule
\textbf{ID} & \textbf{Nominal} & \textbf{Held-out} & \textbf{Success} \\
\midrule
\multicolumn{4}{l}{\textit{\textbf{Adhesive detachment and self-cleaning}}} \\
\addlinespace[1pt]
A1 &
$t_{\mathrm{glue}}\ge 3$ mm; size $50{\times}40{\times}40$ mm &
sizes $\{40{\times}30{\times}30,\,60{\times}50{\times}50\}$; unseen shell &
detach; lift $\ge h_{\mathrm{lift}}$ for $\tau$; no damage \\
A2 &
pad $15{\times}15$ mm; part $65{\times}30{\times}30$ mm &
triangular/quad prism; poses $\pm 30^\circ$ yaw &
detach; clearance for $\tau$; no violent ejection \\
A3 &
patch $150{\times}100{\times}1$ mm; mass $\approx 20$ g &
blue; ring/strip shapes; placement $\pm 50$ mm &
remove; residue $\le 5\%$ by area differencing \\
A4 &
log $200$ mm long, $100$ mm dia.; specimen $50{\times}40$ mm &
log rotated $\pm 45^\circ$; specimen $\pm 30$ mm &
detach; lift $\ge h_{\mathrm{lift}}$ for $\tau$; no damage \\
A5 &
Blu-Tack residue $20{\times}20{\times}5$ mm &
clay; thin-sheet and thick-blob shapes &
fully removed by visual inspection \\
\addlinespace[2pt]
\midrule
\multicolumn{4}{l}{\textit{\textbf{Surface mark removal}}} \\
\addlinespace[1pt]
B1 &
plate dia. $255$ mm; region $\Omega=120{\times}120$ mm &
unseen location/shape/color; green marker &
cleaned area $\ge 90\%$ within $\Omega$ \\
B2 &
hand $90{\times}400{\times}30$ mm; region $\Omega=120{\times}120$ mm &
unseen location/shape/color; blue ink &
cleaned area $\ge 90\%$ within $\Omega$ \\
\addlinespace[2pt]
\midrule
\multicolumn{4}{l}{\textit{\textbf{Occluded retrieval in particulate media}}} \\
\addlinespace[1pt]
C1 &
fossil $80{\times}40{\times}30$ mm; tray $350{\times}500{\times}60$ mm; depth $15$ mm &
unseen fossils $\{80{\times}40{\times}30,\,70{\times}30{\times}30\}$ mm; depths $\{10,\,20\}$ mm; sand camouflage &
exposed $\ge 80\%$; trench depth $\le 20$ mm \\
C2 &
clump $70{\times}15{\times}20$ mm &
color matching; pose $\pm 60$ mm &
clump surface exposed $\ge 80\%$ \\
\bottomrule
\end{tabular}
\end{table*}

\section{Task Details and Success Criteria}
\label{app:task_details}

This appendix summarizes the task parameters and evaluation criteria used for reproducibility. Per-task nominal specifications, held-out conditions, and success criteria are listed in Table~\ref{tab:task_details_physical}, with nominal values matched to Table~\ref{tab:task-preparation}.

\subsection{Evaluation Conventions}
Each rollout terminates at $T_{\max}=120$\,s ($N_{\max}=1800$ steps at $15$\,Hz) and is
successful if the task criterion is met before termination. For sustained-lift tasks,
the object must remain at least $h_{\text{lift}}=10$\,mm above the surface for
$\tau=1.0$\,s. ``No visible damage'' means no visible gouges/tears larger than $2$\,mm.
For cleaning/exposure, the evaluation region $\Omega$ is predefined and percentages are
computed by before/after HSV-space pixel differencing with a fixed threshold.

\subsection{Splits and Scoring}
Training uses $N_{\text{demo}}=20$ demonstrations per task under nominal conditions. When manual scoring is needed, each rollout is labeled by three raters and verified once using the recorded video.

\end{document}